\documentclass{article}

\usepackage[final,nonatbib]{neurips_2024}

\usepackage[utf8]{inputenc} 
\usepackage[T1]{fontenc}    
\usepackage[pagebackref, breaklinks=true, colorlinks,citecolor=citecolor, linkcolor=linkcolor, bookmarks=false]{hyperref}

\usepackage{url}            
\usepackage{booktabs}       
\usepackage{amsfonts}       
\usepackage{nicefrac}       
\usepackage{microtype}      
\usepackage{xcolor}         

\usepackage{subcaption}
\usepackage{nicematrix}
\usepackage{multirow}
\usepackage[most]{tcolorbox}
\usepackage{tabularx}

\def\NAME{Liquid\xspace}
\definecolor{mygray}{gray}{0.8}
\usepackage{subcaption}
\usepackage{nicematrix}
\definecolor{mygray}{gray}{0.8}
\newcommand{\finding}[2]{
    \begin{tcolorbox}[
        colback=white!90!gray,     
        colframe=teal!60!black,     
        arc=5pt,                    
        boxsep=5pt,                 
        left=10pt,                  
        right=10pt,                 
        top=2pt,                    
        bottom=2pt,                 
        boxrule=0.8pt,              
        drop shadow=gray!50!white,  
        enhanced jigsaw             
    ]
    \vspace{-0.1cm}
        \paragraph{\textbf{\textit{Finding #1:}}} #2
    \vspace{-0.1cm}
    \end{tcolorbox}
    \vspace{-0.1cm}
}

\usepackage{amsmath,amsfonts,bm}

\usepackage{pifont}

\def\figref#1{Fig.~\ref{#1}}

\def\eqref#1{(\ref{#1})}

\def\1{\bm{1}}

\DeclareMathAlphabet{\mathsfit}{\encodingdefault}{\sfdefault}{m}{sl}
\SetMathAlphabet{\mathsfit}{bold}{\encodingdefault}{\sfdefault}{bx}{n}

\usepackage{graphicx}
\usepackage{url}            
\usepackage{booktabs}       
\usepackage{nicefrac}       
\usepackage{microtype}      
\usepackage{wrapfig}

\usepackage{enumitem}

\usepackage[ruled,noend,linesnumbered]{algorithm2e}

\usepackage{multirow}
\usepackage{tablefootnote}

\usepackage{color}
\usepackage{colortbl}
\usepackage{xcolor}

\definecolor{blgrey}{rgb}{0.6,0.6,0.6}
\definecolor{bblue}{rgb}{0.855,0.933,0.98}
\definecolor{dblue}{HTML}{5297D6}
\definecolor{gainred}{rgb}{0.1,0.5,0.3}
\definecolor{citecolor}{HTML}{0071BC}
\definecolor{linkcolor}{HTML}{ED1C24}

\usepackage{listings}
\usepackage{color}

\definecolor{dkcyan}{cmyk}{1,0,0,.25}
\definecolor{dkgreen}{rgb}{0,0.6,0}
\definecolor{gray}{rgb}{0.5,0.5,0.5}
\definecolor{mauve}{rgb}{0.58,0,0.82}

\title{Liquid: Language Models are Scalable and Unified \\ Multi-modal Generators}

\vspace{-5pt}
\author{
  \vspace{-25pt}\\
  \textbf{Junfeng Wu$^{1,2}$,\quad Yi Jiang$^{2,\dag}$,\quad Chuofan Ma$^{2,3}$,} \\  \textbf{Yuliang Liu$^{1}$,\quad Hengshuang Zhao$^{3}$, }
  \textbf{ Zehuan Yuan$^{2}$,\quad Song Bai$^{2 , *}$,\quad Xiang Bai$^{1, }$\thanks{Corresponding authors: $<$\href{xbai@hust.edu.cn}{\color{black}{xbai@hust.edu.cn}}$>$, $<$\href{songbai.site@gmail.com}{\color{black}{songbai.site@gmail.com}}$>$; $\dag$: project lead  }
  }\vspace{3pt} \\
  $^1$Huazhong University of Science and Technology \quad $^2$Bytedance Inc  \\ $^3$The University of Hong Kong\vspace{3pt}\\
  \texttt{\small wjf5203@gmail.com, jiangyi.enjoy@bytedance.com,} \\ 
  \texttt{\small b20mcf@connect.hku.hk, ylliu@hust.edu.cn, hszhao@cs.hku.hk}  \\
  \texttt{\small yuanzehuan@bytedance.com, songbai.site@gmail.com, xbai@hust.edu.cn}\vspace{8pt}  \\
  \vspace{-7pt} \\
}

\begin{document}

\maketitle

\begin{abstract}


We present \NAME, an auto-regressive generation paradigm that seamlessly integrates visual comprehension and generation by tokenizing images into discrete codes and learning these code embeddings alongside text tokens within a shared feature space for both vision and language. Unlike previous multimodal large language model (MLLM), \NAME achieves this integration using a single large language model (LLM), eliminating the need for external pretrained visual embeddings such as CLIP. 
For the first time, \NAME uncovers a scaling law that performance drop unavoidably brought by the unified training of visual and language tasks diminishes as the model size increases.
Furthermore, the unified token space
enables visual generation and comprehension tasks to mutually enhance each other, effectively removing the typical interference seen in earlier models. 
We show that existing LLMs can serve as strong foundations for \NAME, saving 100× in training costs while outperforming Chameleon in multimodal capabilities and maintaining language performance comparable to mainstream LLMs like LLAMA2. \NAME also outperforms models like SD v2.1 and SD-XL (FID of 5.47 on MJHQ-30K), excelling in both vision-language and text-only tasks. This work demonstrates that LLMs such as Qwen2.5 and GEMMA2 are powerful multimodal generators, offering a scalable solution for enhancing both vision-language understanding and generation. The code and models are released at \url{https://github.com/FoundationVision/Liquid}.

\end{abstract}
\section{Introduction}
\label{sec:intro}


\begin{figure}[t]
\centering
\includegraphics[width=0.98 \linewidth]{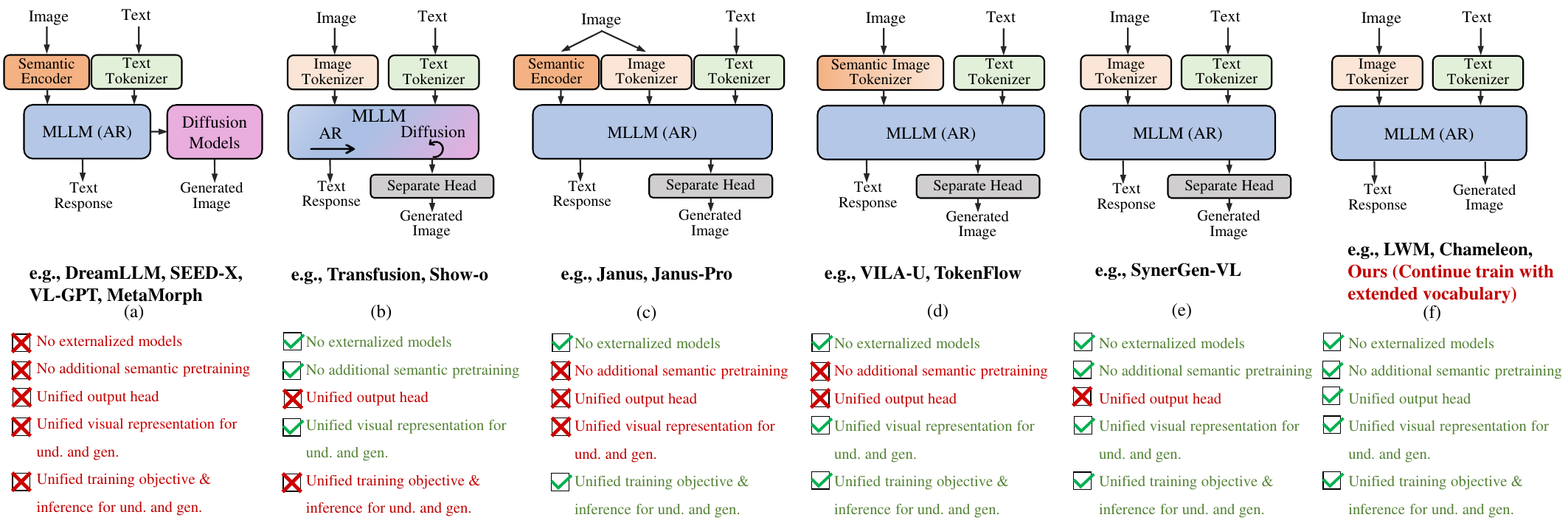}
\caption{
Compared to existing unified MLLMs, \NAME eliminates the need for additional model composition, semantic pre-training, or separate output heads. It directly performs next-token prediction and training within a unified vision-language space, requiring only an extension of vocabulary and low-cost continue training. This design enables seamless adaptation to any existing LLMs.
}
\label{fig:pipeline_compare}
\vspace{-2em}
\end{figure}

The recent advancement of Large Language Models (LLMs)~\cite{gpt3,llama,PALM, T5,touvron2023llama,team2024gemma} has sparked a trend of extending the foundational capabilities of LLMs to the visual domain, giving rise to Multi-modal Large Language Models (MLLMs)~\cite{llava, zhu2023minigpt, liu2024improved, lu2023empirical, chen2024allava, liu2024llavanext}. In the realm of visual understanding, MLLMs such as LLaVA~\cite{llava} typically adopt the pretrained CLIP~\cite{CLIP} model as the visual tokenizer, followed by a two-stage training process to align the vision-and-language feature space. While for text-guided visual generation, most MLLMs~\cite{ge2023planting, ge2024seed, jin2023unified, sun2023generative, sun2024generative, tong2024metamorph} rely on an external diffusion model to generate images as shown in \figref{fig:pipeline_compare}(a). Despite these methods have achieved remarkable multi-modal understanding and generation performances, the use of external visual modules introduces additional architecture complexity to the system, and potentially poses a bottleneck when scaling up the LLMs.

In light of the aforementioned issues, an emerging line of research~\cite{esser2021taming, DALLE, ding2021cogview, yu2022scaling,VAR,sun2024autoregressive} attempts to employ the VQVAE~\cite{van2017neural, esser2021taming} model as a universal visual tokenizer for MLLMs. Analogous to the role played by the BPE~\cite{bpe} tokenizer in LLMs, VQVAE establishes a bi-directional mapping between raw pixels and discrete codes. This enables the MLLMs to learn visual code embeddings jointly with text tokens, rather than constrained by the feature space of a pretrained visual encoders like CLIP or pretrained visual generators like diffusion models. 
Moreover, the discrete nature of visual tokens allows uniform modeling of visual and text tokens with the same next-token prediction loss, which seamlessly integrates both modalities. LWM~\cite{liu2024world} and Chameleon~\cite{team2024chameleon} are the pioneers in exploring this approach. However, these methods involve extensive training from scratch, which makes it computationally expensive to disseminate exploration in this form. Follow-up works further introduce diffusion modeling~\cite{transfusion, xie2024show} but leads to inconsistent training objectives between visual and text generation as show in \figref{fig:pipeline_compare}(b). (c) and (d) require additional semantic pretraining for image tokenizers~\cite{wu2024janus, wu2024vila, qu2024tokenflow}, while (e) necessitates additional preprocessing and postprocessing steps for visual tokens~\cite{li2024synergen}. EMU3~\cite{wang2024emu3} has verified that the next prediction paradigm can achieve SOTA results in both understanding and generation tasks by finetuning separate two models, but there is a lack of exploration of the effects of unifying these two tasks within one model.
In this paper, we explore the potential of solely employing LLMs as multimodal generators, and conduct a comprehensive empirical study to investigate a series of properties including \textbf{the scaling laws of visual generation, the impact on original language capabilities, and the interrelation between visual understanding and generation tasks. }

We first revisit the plain design of MLLMs and present \NAME, a scalable decoder-only architecture for multi-modal generation and understanding. 
We employ the VQGAN~\cite{esser2021taming} as image tokenizer to encode images into discrete tokens, similar to how a BPE processes text. This allows images and text to share the same vocabulary space, enabling LLMs to understand and generate images without any structural modifications.
We find that existing LLMs are excellent starting points for training, as they have already acquired strong semantic understanding and generation capabilities.
Compared to training from scratch like Chameleon~\cite{team2024chameleon}, this approach saves ${100\times}$ of training cost while achieving stronger multimodal capabilities.
We employ language data and image-text pair data to train six different sizes of LLMs, ranging from 0.5B to 32B across various model families. For each model, we create three distinct versions by training with text-only data, image generation data, and a combination of both tasks. This approach allows us to analyze the performance and interrelation between these tasks across different scales.

 Our experiments reveal several insightful properties: 1) Directly employing LLMs for visual generation exhibits \textbf{clear scaling laws} in both validation loss and image consistency metrics, consistent with those seen in LLMs, regardless of whether the models retain their language capabilities or focus solely on visual generation tasks. 2) When trained with multimodal data such as image-text pairs and pure language data, the language capabilities of the models are lower compared to those trained with only language data. However, this \textbf{tradeoff diminishes as the model size increases}, indicating that larger models have sufficient capacity to handle both tasks seamlessly, demonstrating the strong potential of LLMs as unified multimodal generator. 3) \textbf{The visual understanding and visual generation tasks can mutually benefit each other.} We find that increasing the data for either visual generation or visual understanding tasks improves the performance of the other, demonstrating the advantages and natural rationale of this paradigm in joint optimization.

We evaluate the capabilities of \NAME across text-guided image generation, visual understanding, and general text-only tasks. For image generation, \NAME outperforms other auto-regressive based models, as well as some diffusion models like SD-XL and achieve FID of 5.47 on MJHQ-30K, demonstrating that LLMs can acquire excellent imagery capabilities efficiently with a limited amount of data. For visual understanding, \NAME surpasses Chameleon and achieved results comparable to those of well-established MLLMs. In text-only tasks, \NAME achieves comparable performance with Chameleon, which used mix pre-training on a very large scale,  and surpasses the performance of LLAMA2, demonstrating undegraded linguistic capabilities.

In summary, the main contributions of this paper can be categorized into the following points:
\begin{itemize}
    \item {An efficient decoder-only multi-modal generation framework that seamlessly carries out visual generation, visual comprehension, and pure language tasks.}
    \item Comprehensive experiments about scaling laws of unified autoregressive multi-modal models, revealing that the trade-off between language and visual tasks diminishes as scale of model increases.
    \item An insightful discovery regarding the mutual boost of visual understanding and generation tasks within LLMs through the unification of visual tokens, highlighting the potential of LLMs to improve both understanding and generation capabilities by mixtraining.

\end{itemize}


\section{Preliminaries}
\label{sec:Method}

\textbf{Image Tokenizer.}
We use a VQGAN~\cite{esser2021taming} from Chameleon~\cite{team2024chameleon} as image tokenizer. It encodes a 512 × 512 image into 32 × 32 discrete tokens in a codebook of size 8192, which are flattened into a 1024 image token for LLM.
In LLMs such as Llama~\cite{llama} and Gemma~\cite{team2024gemma}, the raw text input will also be tokenized into numbers (discrete IDs) representing the positions of the words in the vocabulary by the BPE~\cite{bpe} tokenizer.
For example, "a small dog" is tokenized into [235250, 2301, 5929] in Gemma, the vocabulary size is 256,000. 
Therefore, we only need to extend the entire vocabulary size to $256,000 + 8192$, where the former for text and the latter for image. 
In this way, both vision and language are represented as sequences of discrete IDs, whether for input or output.
After that, IDs index the input embeddings through an embedding-layer, thus completing the unification of the embedding space. 
Correspondingly, we extend the original LM head by 8192 dimensions to enable the model to predict both text and image tokens within the same embedding space.

\begin{figure*}[t]
\centering
\includegraphics[width=0.98 \linewidth]{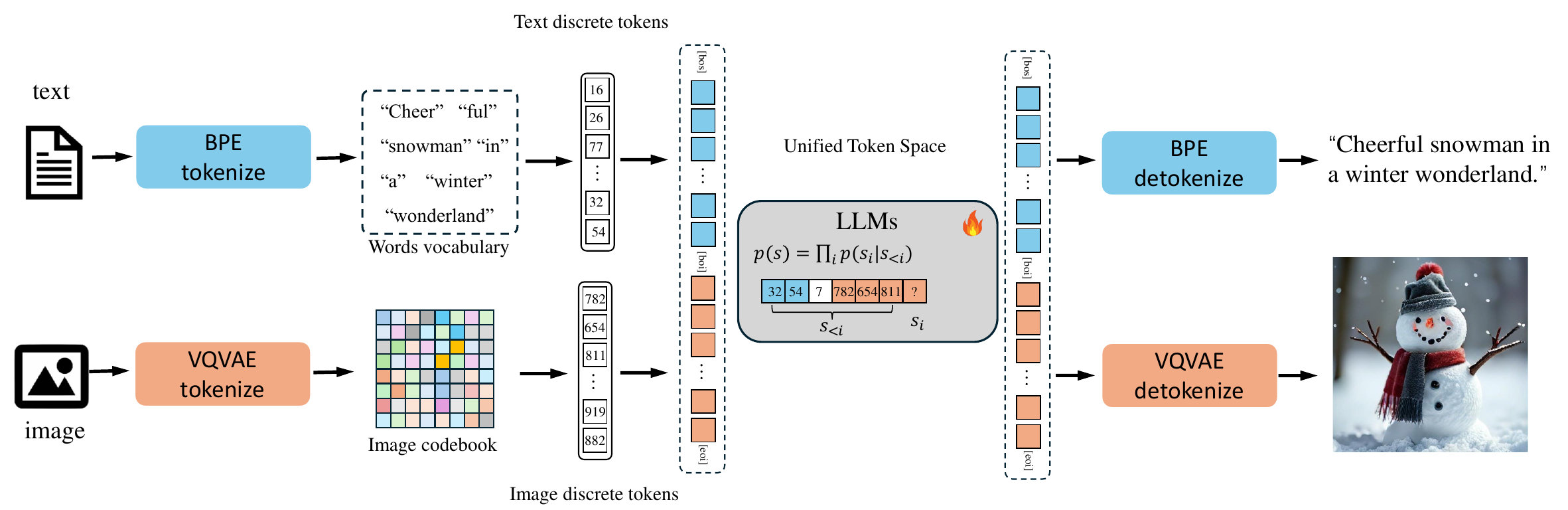}
\caption{
Pipeline of \NAME. The structure of \NAME follows a consistent format that treats images in the exact same way as text. A VQVAE based image tokenizer transforms the input images into discrete codes, which sharing the same vocabulary and embedding space with text codes. The image tokens and text tokens, once mixed, are fed into the LLMs and trained in the form of next token prediction. The lower part of the figure demonstrates that \NAME can handle various multi-modal understanding and generation tasks.
}
\label{fig:pipeline}
\end{figure*}

\textbf{Architecture.}
As illustrated in Fig.~\ref{fig:pipeline}, \NAME can be constructed based on any existing LLMs. In this paper, we employ the GEMMA-7B~\cite{team2024gemma} as base model of \NAME to validate its multimodal understanding, image generation capabilities, and performance on text-only tasks after augmenting it with the ability to understand and generate images. 
To further investigate whether a trade-off exists when accommodating visual generation tasks and text generation tasks within the same LLM space, as well as to explore the scaling performance of different sized models,
we conduct same training on Llama-3 1B ~\cite{dubey2024llama}, GEMMA-2 series~\cite{team2024gemma2} at scales of 2B, 9B, and Qwen2.5 series~\cite{hui2024qwen2} at scales of 0.5B, 7B, 32B, thereby observing their distinct performances. 
We refrain from altering any structures within the LLMs to facilitate continued training directly from pre-trained weights. The only modification is the addition of 8192 new learnable embeddings for discrete image tokens. 
Correspondingly, we extend the original LM head by 8192 dimensions to enable the model to predict both text and image tokens within the same embedding space. 
The LLMs maintain their original next-token prediction training objective with cross-entropy loss. 
, with the only variation being that their vocabulary space has been extended from purely language-space to a combined space for both visual and linguistic modalities.

\textbf{Data Preparation.}
To maintain the language abilities of the pre-trained LLMs, we sample text-only data from public datasets for the pre-training stage. We sample 15M text data from DCLM~\cite{li2024datacomplm}, 12M from SlimPajama~\cite{cerebras2023slimpajama}, and 3M code data from Starcoderdata~\cite{li2023starcoder}, totaling 30M text instances for approximately 60 billion text tokens. 
For image-text pairs, we use JourneyDB~\cite{pan2023journeydb} and internal MidJourney-style synthetic data to compile 30M high-quality image data, for a total of 30 billion image tokens. 
All the data are used to form a hybrid multi-modal data for continue pre-training, resulting in quick acquisition of image generation capabilities while retaining language abilities.

\textbf{Training Procedure.}
We use a total of 60M data to continue pretraining our models. For all image-text pair data, we define the input format for multi-modal training data as:
\[
\small{\text{[bos] \{text token\} [boi] \{image token\} [eoi][eos] }}.
\]
Here, [bos] and [eos] are the begin-of-sequence and end-of-sequence tokens defined in the original text tokenizer. We incorporate two additional spatial tokens, [boi] and [eoi] to signify the start and end of image tokens. Furthermore, we reverse 20\% of text-to-image data as captioning task to enhance the visual understanding abilities.
In the scaling experiments, for each model size, we independently use 30M text-only data, 30M text-to-image data, and a mixed dataset of 60M data to train three different versions of the models. We then evaluate their performance on a series of tasks.

Since all images are tokenized into discrete image tokens that share the same vocabulary and embedding space as text tokens, the training objective is consistent with LLMs via next-token prediction and utilizes the standard cross-entropy loss. Moreover, we observe that during the continue pre-training stage, models of 7B size and above tend to encounter loss spikes in the early stages of training. It significantly impacts the training performance and convergence speed. To mitigate this issue, we reduce the max grad norm to 0.5 for larger models and use max-z loss~\cite{yang2023baichuan} to normalize the logits, enhancing the stability of training.

\textbf{Training Details.}
In all experiments, we initiate with a learning rate of 2e-5. Except for the scaling experiments that employed a constant learning rate, all other tests utilized a cosine learning schedule. The training batch size is set at 1024, and the max context length is set to 2048. We use the DeepSpeed Zero3 optimization strategy.

\section{Scaling, Trade-offs, and Synergy in Unified Multi-modal Generation}
\label{sec:Findings}

In this section, we investigate the following questions under the unified decoder-only multi-modal paradigm of \NAME:

\begin{tabularx}{0.92\linewidth}{@{\hspace{0em}}l@{\hspace{0.5em}}X@{}}
  \S\ref{find_visual_scaling} & Does vision generation adhere to the scaling laws observed in LLMs?  \\
   \S\ref{vision_language_compromise} & Do vision generation and language tasks exhibit mutual interference or enhancement? \\
  \S\ref{gen_und_synergy} & Is there a mutually beneficial relationship between vision understanding and generation?
\end{tabularx}

\begin{figure*}[tb]
\centering
\includegraphics[width=1.0 \linewidth]{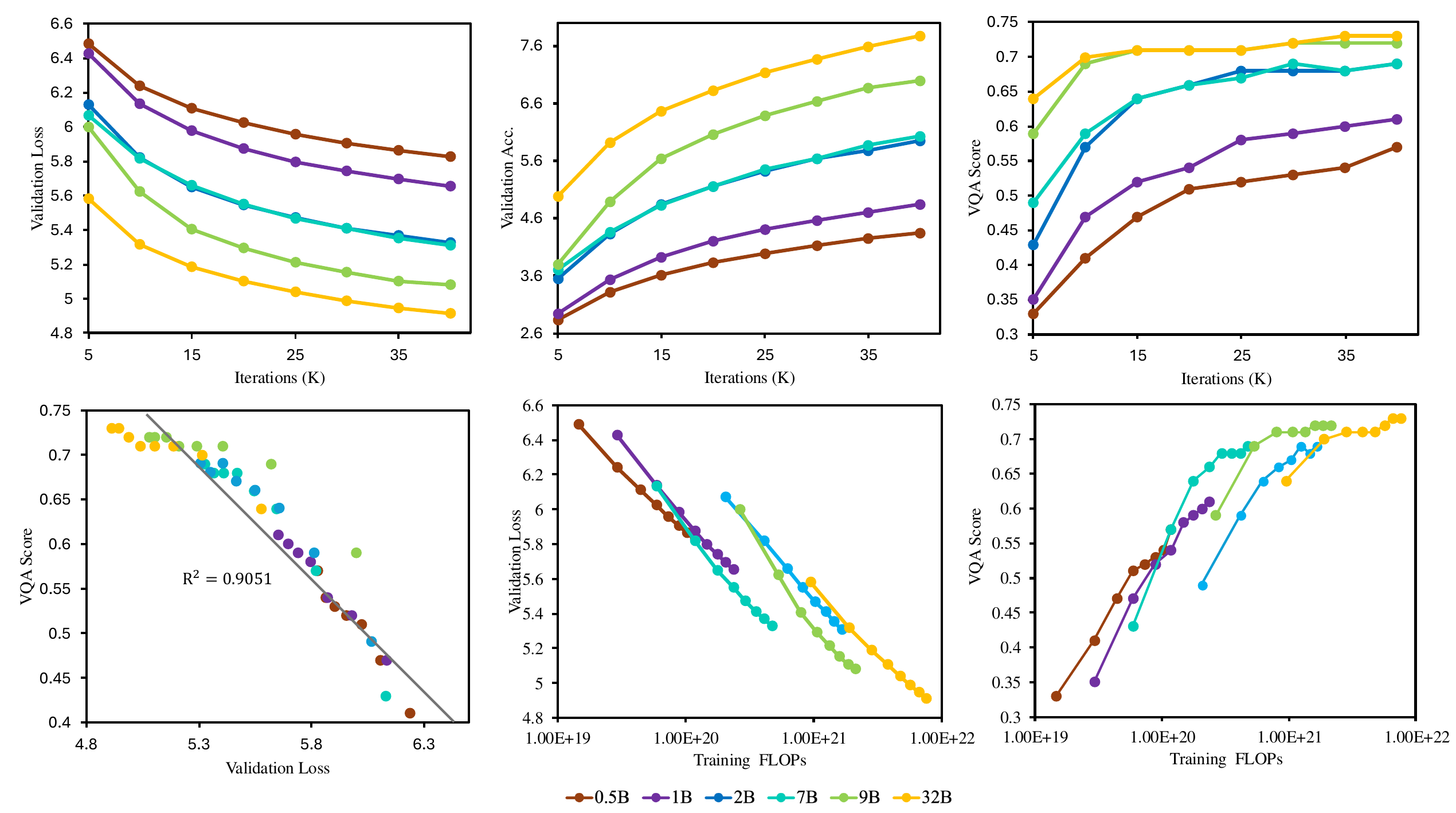}
\caption{
\textbf{Scaling behavior of LLMs in multimodal generation across different sizes.} We explore the visual generation performance of LLMs ranging from 0.5B to 32B in size after undergoing mixed training with language data and text-to-image data. The validation loss smoothly decreases as both the model size and training iterations increase, while token accuracy and VQA Score consistently rise. The VQA Score consistently increases as the validation loss decreases, indicating a strong correlation between them. Under the same training FLOPs, smaller models reach lower validation losses more quickly, but larger models ultimately achieve higher evaluation metrics.
}
\label{fig:mixpretrain_t2i_scaling}
\end{figure*}

\textbf{Evaluation settings.} We use validation loss, token accuracy, and VQAscore from GenAI-Bench~\cite{GenAI-Bench} to evaluate the performance of visual generation tasks. For language ability, we use the average score across five widely-used benchmarks: 
HellaSwag~\cite{zellers2019hellaswag}, WinoGrande~\cite{sakaguchi2021winogrande} , ARC-Easy~\cite{clark2018think}, ARC-Challenge~\cite{clark2018think}, and BoolQ~\cite{clark2019boolq}.
For visual understanding tasks, we conduct evaluations on VQA-v2~\cite{vqa_v2}, GQA~\cite{gqa}, TextVQA~\cite{textvqa}, POPE~\cite{POPE}, and MME~\cite{mme}.

\subsection{Scaling Results on Visual Generation}
\label{find_visual_scaling}
\begin{figure*}[tb]
\centering
\includegraphics[width=0.9 \linewidth]{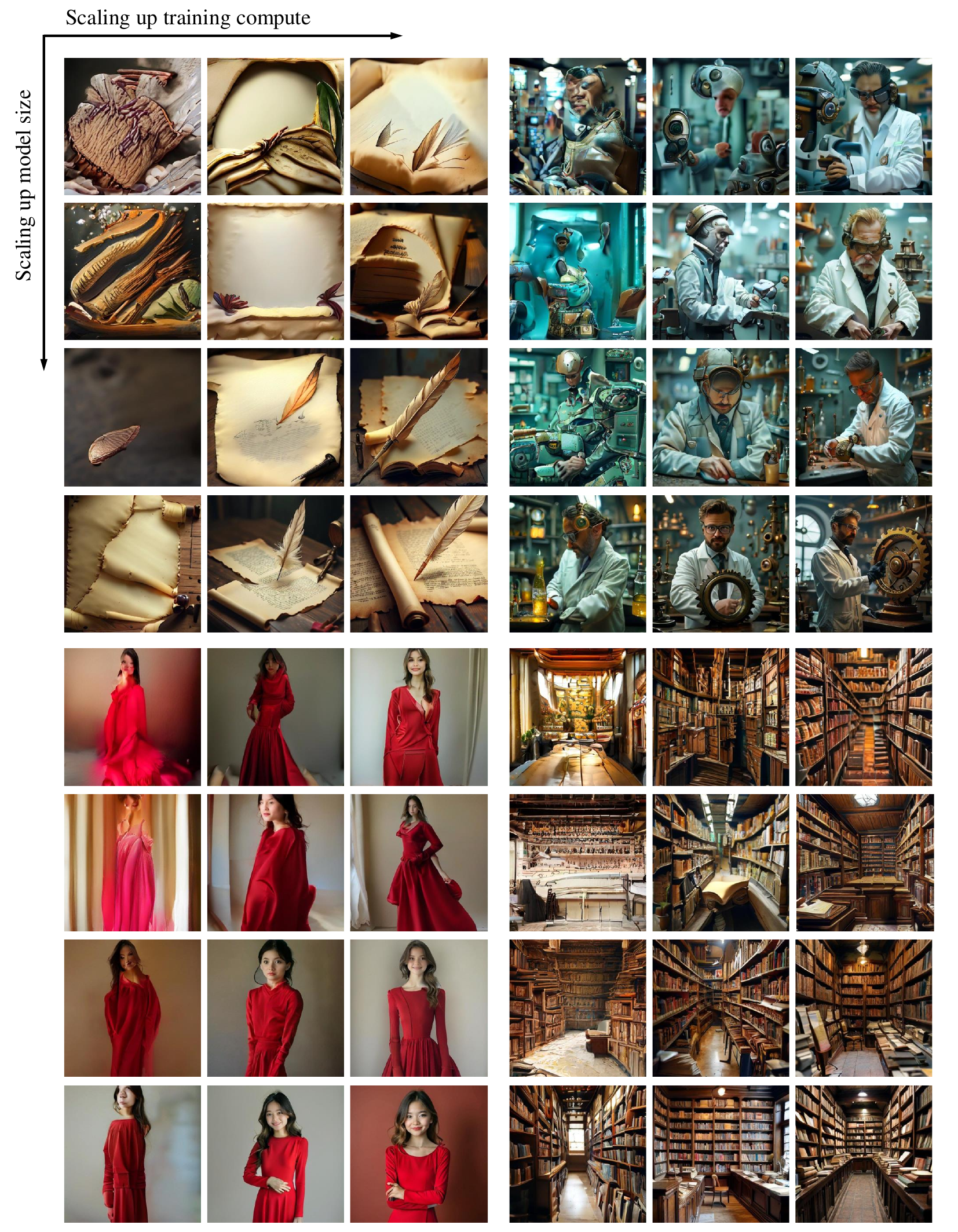}
\caption{Scaling model size and training compute improves visual fidelity and soundness. Zoom in for a better view. Samples are drawn from Liquid models of 4 different sizes (0.5B, 1B, 2B, 9B) and 3 different training steps (5K, 15K, 40K).}
\label{fig:mixpretrain_t2i_scaling_visualization}
\end{figure*}

We explore the visual generation performance of LLMs ranging from 0.5B to 32B in size after mixed training with 60M language and text-to-image data. As shown in Fig.~\ref{fig:mixpretrain_t2i_scaling}, with the increase in model size and training iterations, the validation loss smoothly decreases, while token accuracy and VQA Score consistently increase. The VQA score continues to rise as the validation loss decreases, indicating a strong correlation between validation loss and holistic image evaluation metrics. 
Moreover, we observe a clear power-law scaling trend for Loss as a function of Training FLOPs, as consistent with LLMs.
Under the same training FLOPs, we find smaller models achieve lower validation losses and higher VQA Scores more quickly, but larger models ultimately reach higher evaluation metrics. This may be attributed to the fact that smaller models can train for more steps under the same FLOPs, enabling them to transition more quickly from the language domain to the multimodal domain.
Although smaller models can quickly acquire visual generation abilities, their upper limits are lower, making it difficult to achieve high-quality visual generation results. 
For further visualizations on scaling model size and training compute, please refer to Fig.~\ref{fig:mixpretrain_t2i_scaling_visualization}.

\finding{1}{After equipping LLMs with visual capabilities through vocabulary expansion, their performance on visual generation tasks still follows the power-law scaling law observed in language tasks.}

\begin{figure*}[h]
\centering
\includegraphics[width=0.8 \linewidth]{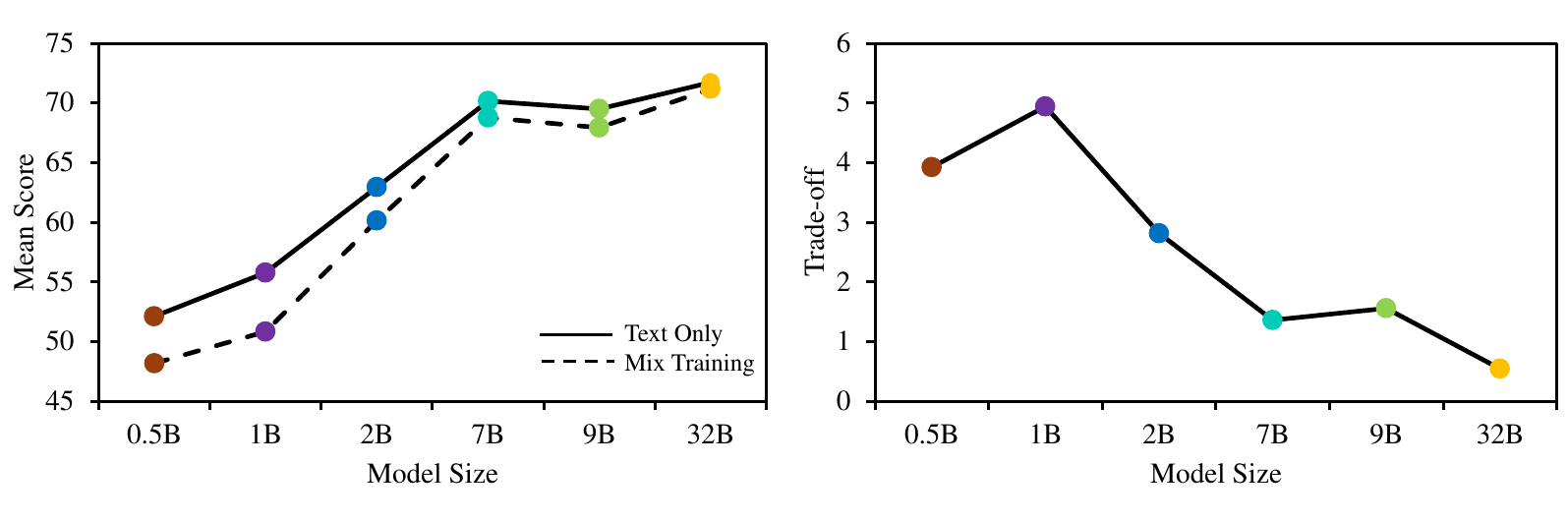}
\vspace{-4ex}
\caption{
The performance impact of multi-modal mixed training versus text-only training on language tasks. Multi-modal mixed training does impact language performance when the model size is small. However, this degradation gradually disappears as the model size increases.
}
\vspace{-2ex}
\label{fig:text_tradeoff}
\end{figure*}

\begin{figure*}[h]
\centering
\includegraphics[width=0.95 \linewidth]{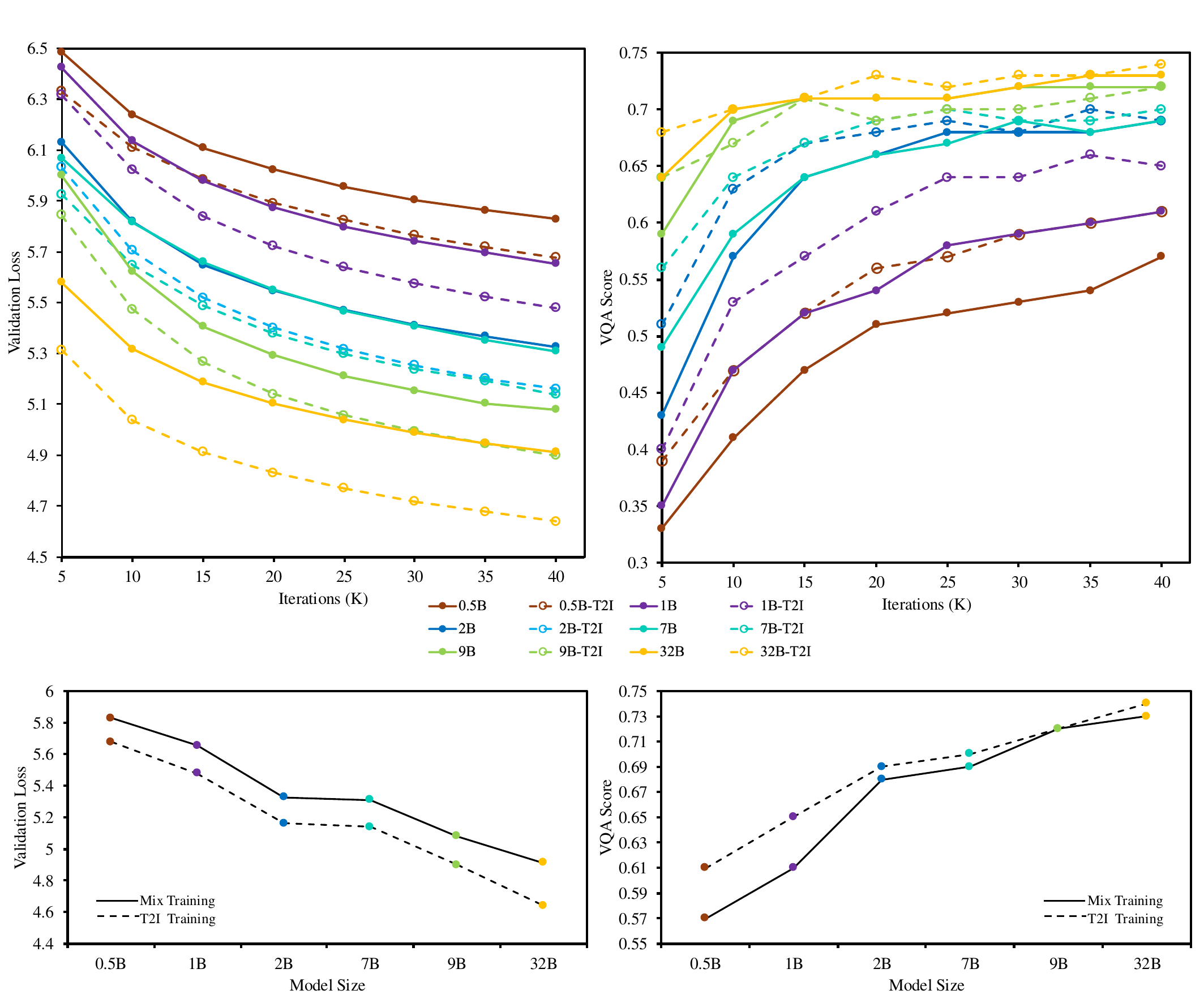}
\vspace{-2ex}
\caption{Scaling results between T2I only training and Mix training. Mix training results in higher validation loss on visual generation tasks for models of every size. However, its impact on the VQA Score diminishes as the size of the model increases.}
\label{fig:t2i_mix_scaling}
\end{figure*}

\subsection{Is there a conflict between visual and language generation?}
\label{vision_language_compromise}

To explore the mutual influence between visual and language generation, we train models of six different sizes under three distinct settings and compared their performances. In addition to training on a mixed multimodal dataset of 60M mix data, we separately train two sets of models, one on 30M text-only data and the other on 30M text-to-image data, to isolate language and visual generation capabilities. We then compare the performance gaps between these isolated models and the mixed model on both language tasks and visual generation tasks.

To investigate the impact on language capabilities, we evaluate the language performance of the mixed pre-train model against the model pre-trained solely on text-only data, as illustrated in Fig.~\ref{fig:text_tradeoff}, we find a trade-off phenomenon when both tasks were mixed in the training of smaller models; however, as the model size increased, this trade-off gradually disappear. It confirms that larger models possess sufficient capacity to handle both visual and language spaces generation concurrently.

In language tasks, we observe a trade-off in performance of multimodal LLMs, raising the question of whether this trade-off also exists in visual generation tasks. 
As illustrated in Fig.~\ref{fig:t2i_mix_scaling}, mixed training leads to higher validation loss on visual generation tasks for models of every size. However, the detrimental effect on the VQA-Score decreases as the model size increases. This reduction is partly due to the increased capacity of larger models and may also be attributed to the evaluation metrics approaching saturation.
Nevertheless, it can be argued that as the size of the model increases, the trade-off between visual generation and language tasks gradually diminishes or even disappears. This encourages leveraging the scaling capabilities of LLMs into multimodal models.

\finding{2}{There exists a trade-off between visual and language generation, but this trade-off gradually diminishes as the model size increases.}

\begin{table*}[tb]
    \resizebox{1.0\linewidth}{!}{
    \begin{tabular}{lccc|cccc|ccccc} 
    \toprule 
    \multirow{2}{*}{\bf Method}   & \multicolumn{3}{c}{ {\it{\bf Training Data }}}     & \multicolumn{3}{c}{ {\it{\bf Visual Generation}}}     & \multicolumn{5}{c}{ {\it{\bf Visual Understanding}}}      \\
     \cmidrule(lr){2-4} \cmidrule(lr){5-7}   \cmidrule(lr){8-12}  
    &Text-only &Visual Gen. &Visual Und. &gFID$\downarrow$ &Basic &Advanced &Overall & { VQAv2} & { GQA} & { TextVQA} & { POPE} & { MME} \\
     
    \hline
    Baseline    &10M&10M&10M  &19.9 &0.63 &0.58 &0.60 &60.7 &51.3 &39.2 &75.9 &909.6 \\
    Add Gen. &10M&\textbf{20M}&10M &12.8  &0.78 &0.63 &0.69 &64.5 &53.2&39.8&78.0&1066.8 \\
    Add Und. &10M&10M&\textbf{20M} &14.9  &0.73 &0.62 &0.66  &63.5 &53.7& 40.5& 76.8 &1035.1\\
    
    \bottomrule
    \end{tabular}
    }
    \centering
        \caption{The impact between visual understanding and generation tasks. ``Visual Gen." refers to the data used for training text-guided image generation, while``Visual Und." refers to the data used for training visual understanding capabilities. Compared to the baseline, adding more understanding data improve the generation capability, and increasing the generation data also aids in understanding ability.
    }
    \label{tab:booster}
\end{table*}

\begin{figure*}[t]
\centering
\includegraphics[width=0.95 \linewidth]{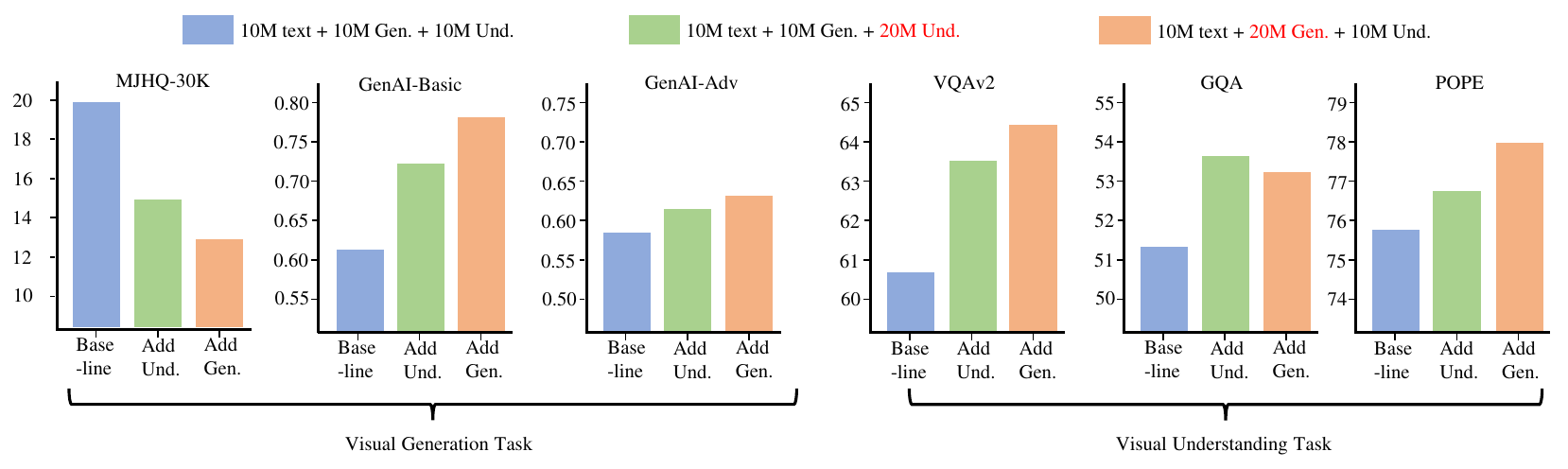}
\caption{Synergistic mutual enhancement between visual understanding and generation tasks: expanding training data for one improves performance in both.}
\label{fig:synergy}
\end{figure*}

\subsection{Will Understanding and Generation Tasks Mutually Improve Each Other?}
\label{gen_und_synergy}

When visual inputs and outputs are modeled by LLMs in the same manner, the gap between them effectively vanishes for LLMs. In this section, we explore whether there exists a synergistic relationship between visual generation and language capabilities.
We conduct three experiments. In the first set, we used a combination of 10M text-only data, 10M visual generation data, and 10M visual understanding data, resulting in a total of 30M data for the pre-training phase, which serves as the baseline. Next, we add 10M visual generation data (`Add Gen.') and 10M visual understanding data (`Add Und.') respectively. After the three pret-raining phases, we fine-tune the models on instruction data and subsequently evaluate them on both visual understanding and generation benchmarks.

As shown in Tab.~\ref{tab:booster} and Fig.~\ref{fig:synergy}, we observe that adding more visual understanding data during pre-training significantly improves the performance of visual generation tasks. Conversely, adding more generation data further enhances the performance of understanding tasks. This phenomenon indicates that when the modality spaces for visual understanding and generation are unified, the training of these two tasks can mutually benefit each other. Intuitively, when images are represented in the same feature space, both visual generation and visual understanding tasks require stronger consistency constraints and sufficient interaction between language and visual information. Therefore, the optimization directions for these two tasks are very similar. It further demonstrates the potential of LLMs as general-purpose multi-modal generators.

\finding{3}{There exists a synergistic relationship between visual understanding and generation, where increasing data for either leads to simultaneous improvements in both.}

\section{Experiments}
\label{sec:Exps}
We will obtain two types of models: the Pretrain model and the Instruction Tuning model (IT). When we initialize with an existing LLM and pretrain it on a mixed dataset of 60M multimodal data, the resulting model is referred to as the Pretrain model. This model retains fundamental language capabilities while additionally acquiring visual generation abilities. We evaluate its performance on both visual generation tasks and language tasks. For visual understanding tasks, we followed the pipeline commonly used in most visual understanding works. Specifically, we performed additional supervised fine-tuning (SFT) on the Pretrain model to enhance its instruction-following capabilities and then assessed its visual understanding performance.

\begin{table}[t]
\setlength{\tabcolsep}{5.5pt}
\centering
\vspace{2ex}
\scalebox{0.8}{
    \begin{NiceTabular}{lccccccc|c}
    \CodeBefore
    \Body
    \toprule[1.2pt]
    \multirow{2}{*}{\textbf{Method}} & \multirow{2}{*}{\textbf{Type}} & \multirow{2}{*}{\#\textbf{Training Images}} & \multirow{2}{*}{\bf Attribute{$\uparrow$}} & \multirow{2}{*}{\bf Scene{$\uparrow$}} & \multicolumn{3}{c}{\bf Relation{$\uparrow$}} & \multirow{2}{*}{\bf Overall$\uparrow$}   \\
    \cmidrule{6-8}
    &  &  & & & Spatial & Action & Part   \\
    \midrule
    SD v2.1 \cite{ldm} & Diffusion & 2000M & 0.80 & 0.79 & 0.76 & 0.77 & 0.80 & 0.78 \\
    SD-XL \cite{sdxl} & Diffusion & 2000M & 0.84 & 0.84 & 0.82 & 0.83 & 0.89 & 0.83\\
    Midjourney v6 \cite{midjourney} & Diffusion & -- & 0.88 & 0.87 & 0.87 & 0.87 & 0.91 & 0.87 \\
    DALL-E 3 \cite{dalle-3} & Diffusion & -- & 0.91 & 0.90 & 0.92 & 0.89 & 0.91 & 0.90 \\
    \midrule
    Show-o \cite{xie2024show} & Discrete Diff.  & 36M & 0.72 & 0.72 & 0.70 & 0.70 & 0.75 & 0.70  \\
    LWM \cite{liu2024world} & Autoregressive & -- & 0.63 & 0.62 & 0.65 & 0.63 & 0.70 & 0.63 \\
   VILA-U~\cite{wu2024vila} (256) & Autoregressive & 15M & 0.78 & 0.78 & 0.77 & 0.78 & 0.79 & 0.76 \\
     VILA-U~\cite{wu2024vila} (384) & Autoregressive & 15M & 0.75 & 0.76 & 0.75  & 0.73  & 0.75 & 0.73 \\
    \midrule
    Liquid-7B & Autoregressive & 30M  &0.84 &0.86 &0.81 &0.83 &0.91 &0.83 \\
    \bottomrule[1.2pt]
    \end{NiceTabular}}\\
    \vspace{1mm}
    (a) VQAScores on \textit{basic} prompts of GenAI-Bench \\
    \vspace{1mm}
\scalebox{0.76}{
    \begin{NiceTabular}{lccccccc|c}
    \CodeBefore
    \Body
    \toprule[1.2pt]
    \multirow{2}{*}{\textbf{Method}} & \multirow{2}{*}{\textbf{Type}} & 
    \multirow{2}{*}{\#\textbf{Training Images}} & 
    \multirow{2}{*}{\bf Count{$\uparrow$}} & \multirow{2}{*}{\bf Differ{$\uparrow$}} & \multirow{2}{*}{\bf Compare{$\uparrow$}} & \multicolumn{2}{c}{\bf Logical{$\uparrow$}} & \multirow{2}{*}{\bf Overall{$\uparrow$}}   \\
    \cmidrule{7-8}
    &  &  & & &  & Negate & Universal   \\
    \midrule
    SD v2.1 \cite{ldm} & Diffusion & 2000M & 0.68 & 0.70 & 0.68 & 0.54 & 0.64 & 0.62 \\
    SD-XL \cite{sdxl} & Diffusion & 2000M & 0.71 & 0.73 & 0.69 & 0.50 & 0.66 & 0.63 \\
    Midjourney v6 \cite{midjourney} & Diffusion & -- & 0.78 & 0.78 & 0.79 & 0.50 & 0.76 & 0.69 \\
    DALL-E 3 \cite{dalle-3} & Diffusion & --
 & 0.82 & 0.78 & 0.82 & 0.48 & 0.80 & 0.70 \\
     \midrule

 Show-o \cite{xie2024show} & Discrete Diff. & 36M & 0.70  & 0.62 & 0.71 & 0.51 & 0.65 & 0.60 \\
    LWM \cite{liu2024world} & Autoregressive & --
 &0.59 & 0.58 & 0.54 & 0.49 & 0.52 & 0.53 \\
    VILA-U~\cite{wu2024vila} (256) & Autoregressive & 15M & 0.70 & 0.71 & 0.74 & 0.53 & 0.66 & 0.64 \\
    VILA-U~\cite{wu2024vila} (384) & Autoregressive & 15M & 0.68 & 0.67 & 0.71  & 0.51  & 0.64 & 0.61 \\
    \midrule
    Liquid-7B & Autoregressive & 30M   &0.76 &0.73 &0.74 &0.46 &0.74 &0.65     \\
    \bottomrule[1.2pt]
    \end{NiceTabular}} \\
    \vspace{1mm}
     (b) VQAScores on \textit{advanced} prompts of GenAI-Bench
        \vspace{2ex}
\caption{Comparison of VQAscore with other visual generation methods on GenAI-Bench.
The basic prompts primarily focus on aspects such as scene, attribute, and relation, while the advanced prompts place a greater emphasis on counting, comparison, differentiation, and logic. 
The advanced prompts require complex visio-linguistic reasoning and present a significantly higher level of difficulty. \NAME outperformes all auto-regressive unified MLLMs on both types of prompts and has even surpassed some well-established diffusion models like SD v2.1~\cite{ldm} and SD-XL~\cite{sdxl}. It demonstrates that the images generated by \NAME align well with the input text prompts.
}
\label{tab:genai_bench}
\vspace{-5mm}
\end{table}

\subsection{Quantitative Results on Visual Generation}

For the image generation tasks, we evaluate the model on three benchmarks: GenAI-Bench~\cite{GenAI-Bench}, MJHQ-30K~\cite{playgroundv2.5}, and WISE~\cite{niu2025wise}.
GenAI-Bench~\cite{GenAI-Bench} is an challenging image-to-text generation benchmark designed to evaluate the capabilities of visual generation models. 
It employs VQAScore, which leverages a visual-question-answering (VQA) model. This enables more precise evaluation of how well the generated image aligns with the text prompt, critically assessing the capability to parse scenes, objects, attributes, relationships, and engage in higher-order reasoning such as comparison and logic. MJHQ~\cite{playgroundv2.5} calculates the Frechet Inception Distance (FID)~\cite{fid} score between the generated images and 30K high-quality images to assess the quality of the generated images.

\begin{wraptable}{r}{7cm}
\centering
\vspace{-2ex}
\resizebox{1.0\linewidth}{!}{
\setlength{\tabcolsep}{2pt}
\begin{tabular}{lccc}
\toprule 
\textbf{Method} & \textbf{Type} & \textbf{\#Images} & \textbf{FID}$\downarrow$ \\
    \midrule
    SD-XL \cite{sdxl} & Diffusion & 2000M & 9.55 \\
    PixArt \cite{chen2023pixart} & Diffusion & 25M & 6.14 \\ 
    Playground v2.5 \cite{playgroundv2.5} & Diffusion & -- & 4.48 \\ 
    Show-o \cite{xie2024show} & Discrete Diff. & 36M & 15.18 \\
    LWM \cite{liu2024world} & Autoregressive & -- & 17.77 \\
    VILA-U (256)~\cite{wu2024vila} & Autoregressive & 15M & 12.81 \\
    VILA-U (384)~\cite{wu2024vila} & Autoregressive & 15M & 7.69 \\
    Janus~\cite{wu2024janus}  & Autoregressive &-  & 10.10 \\
    \midrule
    Liquid-7B & Autoregressive & 30M & 5.47 \\
    \bottomrule
\end{tabular}}
\caption{ Comparison with other visual generation methods on MJHQ-30K evaluation benchmark.
The FID of \NAME is lower than that of all the auto-regressive models and even outperforms most diffusion models. It indicates that the images generated by \NAME have superior aesthetic quality.
}
\label{table:mjhq}
\vspace{-1em}
\end{wraptable}
\looseness=-1
 
WISE~\cite{niu2025wise} is the first comprehensive benchmark designed to evaluate world knowledge integration in text-to-image generation, featuring 1,000 carefully curated prompts spanning 25 diverse subdomains. It proposes WiScore as evaluation metric to systematically assesses knowledge-image alignment through a weighted combination of semantic consistency, physical realism, and aesthetic quality.
These three benchmarks collectively evaluate the generated images from three perspectives: text-image alignment, image realism and fidelity, and world knowledge with reasoning.
 
\begin{figure*}[tb]
\centering
\includegraphics[width=1.0 \linewidth]{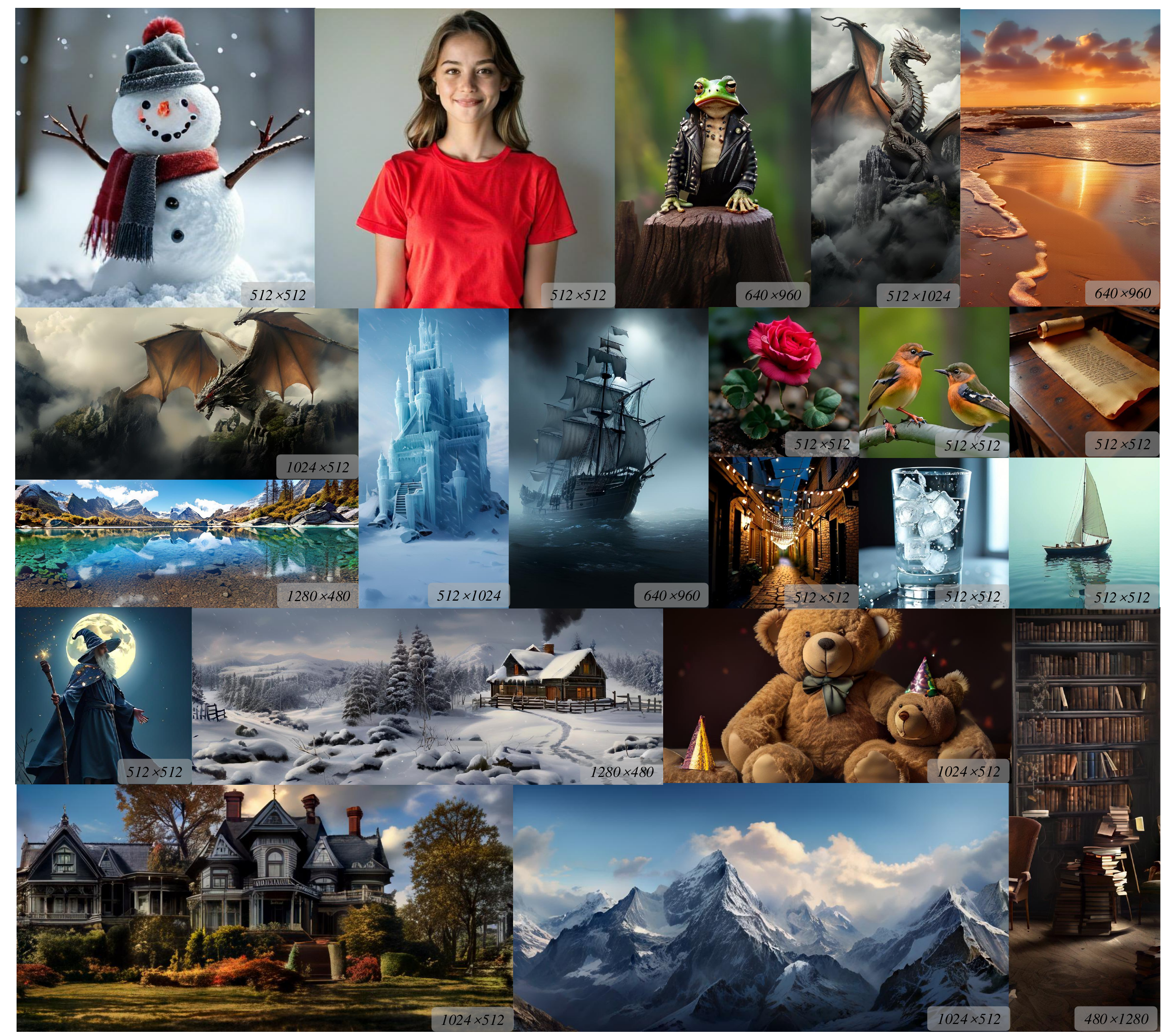}
\caption{The generated samples from \NAME-7B, showcase excellent capabilities in crafting high aesthetic and described-consistent images.}
\label{fig:samples}
\end{figure*}

\begin{table*}[h!]
    \centering
    \resizebox{1.0\linewidth}{!}{
    \begin{tabular}{cc|cccccc|c} 
    \toprule
    Type &Model & Cultural  & Time     & Space    & Biology    & Physics    & Chemistry & \textbf{Overall}   \\
    \midrule
    \multirow{8}{*}{Gen. Only} &FLUX.1-dev &0.48  & \textbf{0.58} &\textbf{0.62 } &0.42  &0.51 & \textbf{0.35 }&\textbf{ 0.50} \\
    &FLUX.1-schnell &0.39  &0.44  &0.50 & 0.31&0.44  &0.26  & 0.40 \\
    &PixArt-Alpha &0.45  & 0.50& 0.48 & \textbf{ 0.49}&\textbf{0.56} &0.34 &  0.47\\
    &playground-v2.5 &\textbf{0.49 } &0.58  & 0.55&0.43  & 0.48&0.33 & 0.49 \\
    &SD-v1-5& 0.34 & 0.35& 0.32&0.28 &0.29 &0.21 &  0.32\\
    &SD-2-1 & 0.30 & 0.38 &0.35 & 0.33 & 0.34&0.21 & 0.32 \\
    &SD-XL-base-0.9 &0.43  & 0.48 &0.47  &0.44  &0.45 &0.27 & 0.43 \\
    &SD-3-medium &0.42  & 0.44 &0.48 &0.39  &0.47 &0.29 & 0.42 \\
    \midrule
    \multirow{10}{*}{Und. \& Gen.} &\textbf{Ours} &\textbf{0.38} &0.42 &\textbf{0.53} &0.36 &\textbf{0.47} &\textbf{0.30} &\textbf{0.41} \\
    &Emu3$^*$ & {0.34}&\textbf{0.45 }&0.48 &\textbf{0.41}  &0.45 &0.27 & {0.39} \\
    &Janus-1.3B &0.16 &0.26 &0.35 & 0.28 &0.30 & 0.14&  0.23\\
    &JanusFlow-1.3B &0.13 &0.26 &0.28 &  0.20& 0.19&0.11 &  0.18\\
    &Janus-Pro-1B & 0.20& 0.28&0.45 & 0.24 & 0.32& 0.16&  0.26\\
    &Janus-Pro-7B & 0.30& 0.37& {0.49}&  0.36&0.42 &0.26 & 0.35 \\
    &Orthus-7B-base &0.07 &0.10 &0.12 & 0.15 &0.15 & 0.10&0.10  \\
    &Orthus-7B-instruct &0.23 &0.31 &0.38 &0.28  & 0.31&0.20 &  0.27\\
    &Show-o-demo & 0.28 &0.36  &0.40&  0.23& 0.33& 0.22 & 0.30 \\
    &VILA-U-7b-256 & 0.26 &0.33  & 0.37 &0.35  &0.39 &0.23 &  0.31\\
    \bottomrule
    \end{tabular}}
    \caption{Normalized WiScore of different models on WISE benchmark. Liquid substantially surpasses all unified MLLMs while rivaling leading specialized visual generation models.
    * indicates the model underwent text-to-image fine-tuning rather than retaining full comprehension and generation capabilities.
    “Und.” and “Gen.” denote “understanding” and “generation”.
    }
    \label{tab:WiScore}
\end{table*}

\textbf{Text-image Alignment}. As shown in Tab.~\ref{tab:genai_bench}, compare with other auto-regressive based methods, \NAME achieves a better overall score under both basic prompts and advanced prompts. This suggests that the images generated by \NAME align better semantically with the input text prompts. 
Notably, \NAME also outperforms some well-established diffusion models like SD v2.1~\cite{ldm} and SD-XL~\cite{sdxl} for both basic and advanced prompts. 
Compared to these diffusion models, \NAME utilizes significantly fewer image data, which indicating that learning based on LLMs can assist the model in understanding the semantic association between the generated content and prompts, while also offering higher training efficiency. Moreover, it demonstrates that LLMs have strong potential for generating complex visual content.


\textbf{Image Realism and Fidelity.} In Tab.~\ref{table:mjhq}, we report FID on MJHQ-30K to compare the images quality generated by \NAME with other models. It is observable that \NAME not only has a lower FID than all other auto-regressive methods but also surpasses most well-known diffusion models except Playground v2.5~\cite{playgroundv2.5}, achieving a very low FID of 5.47. It indicates that LLMs are also capable of generating high-quality images, providing proof that the upper limit of LLMs in terms of image aesthetic quality is not inferior to diffusion models. Further more, due to the capability of LLM to generate dynamically-length content in the form of next-token prediction, the convenience can be applied to visual generation. We find that by appending instructions about the resolution to the input text prompt, such as "length is: width is:", the model can quickly learn to generate the corresponding code according to the specified number of rows and columns. 
Fig~\ref{fig:samples} demonstrates the generation results at various resolutions, showcasing the flexibility of \NAME.

\textbf{World Knowledge and Reasoning.}
Compared to prior methods that only evaluate shallow text-image alignment, WISE introduces more challenging reasoning-driven prompts (e.g., ‘Einstein’s favorite musical instrument’), requiring models to generate images based on deeper semantic understanding. AAs shown in Tab.~\ref{tab:WiScore}, Liquid significantly outperforms other unified MLLMs in complex reasoning scenarios for visual generation and achieves overall WiScore at 0.41, Notably, it even surpasses specialized generative models such as SD series and FLUX.1-schnell, demonstrating its strong retention of world knowledge and reasoning capabilities.

\begin{table*}[t!]
    \centering
    \resizebox{1.0\linewidth}{!}{
    \begin{tabular}{llcccccccccccc}
    \toprule
     Method &BaseModel &Training & BoolQ & PIQA & SIQA & \hspace{-0.1cm} HellaSwag \hspace{-0.1cm} & \hspace{-0.1cm} WinoGrande \hspace{-0.1cm} & ARC-e & ARC-c & MMLU \\
      \midrule
      \multirow{2}{*}{Llama2~\cite{touvron2023llama}}
         & Llama-7B &-- & 77.4 & 78.8  & 48.3  & 77.2  & 69.2 & 75.2 & 45.9   & 45.3\\
         & Llama-13B &-- & 81.7 & 80.5 & 50.3  & 80.7 & 72.8 & 77.3 & 49.4   & 54.8\\
      \midrule  
        \multirow{2}{*}{Chameleon~\cite{team2024chameleon}}
        & Chameleon-7B &-- &81.4 &79.6  &57.0 &74.2 &70.4 &76.1 &46.5 &52.1  \\
        & Chameleon-34B  &-- &86.0 &83.3 &63.3 &82.7 &78.5  &84.1 &59.7 &65.8 \\
      \midrule  
         {Gemma~\cite{team2024gemma}}
        & Gemma-7B  &-- &83.4 &81.2 &51.8 &80.9 &75.1 &53.8 &81.0 &62.6  \\
      \midrule   
      VILA-U~\cite{wu2024vila} & Llama2-7B &multi-modal &70.6 &73.5 &47.7 &54.2 &57.3 &51.6 &34.0 &25.5\\
      \midrule   
        \multirow{8}{*}{\textbf{Ours}}  & Gemma-7B &text-only  &82.1 &81.4 &46.9 &77.2 &71.7 &76.3 &48.8 &55.5  \\
        & Gemma-7B &multi-modal &81.0 &81.0 &46.7 &76.1 &72.7 &75.6 &49.0 &56.0 \\
     &Llama3-1B &text-only &61.0&75.3&41.7 &62.7 &59.7 &61.5 &34.0 &27.2 \\
      &Llama3-1B &multi-modal &53.3 &73.8&40.3 &55.0 &54.5 &59.0 &32.4 &24.5 \\
      &Qwen2.5-7B &text-only &80.8 &79.5 &45.4 &75.4 &70.0 &74.1  &50.4 &64.9  \\
      &Qwen2.5-7B &multi-modal &80.0 &78.7 &45.5 &73.5 &66.7 &74.5 &49.2 &62.0   \\
      &Qwen2.5-32B &text-only &85.2 &80.5 &46.1 &79.2 &74.0 &78.9 &56.1  &68.0  \\
      &Qwen2.5-32B &multi-modal &84.9 &79.8 &45.6 &79.7 &73.5 &78.2  &56.3  &66.5  \\
      \bottomrule
    \end{tabular}}
    \caption{Performance of pre-trained model on standard text-only benchmarks. \NAME outperforms the well-established language model Llama2 and the mix-pretrained multi-modal language model Chameleon in most tasks, exhibiting undegraded linguistic capabilities.}
    \label{table:language_task}
\end{table*}

\subsection{Comparison with Mainstream LLMs}
To validate whether acquiring image understanding and generation capabilities has any impact on the original language abilities of the LLMs, 
we compare our mixed multimodal pretrained model against other state-of-the-art large language models and multi-modal language models across a suite of popular benchmarks that measure commonsense reasoning and reading comprehension capabilities: HellaSwag~\cite{zellers2019hellaswag}, WinoGrande~\cite{sakaguchi2021winogrande} , ARC-Easy~\cite{clark2018think}, ARC-Challenge~\cite{clark2018think}, OpenBookQA~\cite{mihaylov2018can}, PIQA \cite{bisk2020piqa}, SIQA \cite{sap2019socialiqa}, and BoolQ~\cite{clark2019boolq}. We also perform an evaluation of the 5-shot results on MMLU~\cite{Hendrycks2020MeasuringMM}, a comprehensive benchmark that measures world/in-domain knowledge and problem-solving skills across 57 subjects. 

As shown in Tab.~\ref{table:language_task}, \NAME outperforms the well-established language model Llama2~\cite{touvron2023llama} and the mix-pretrained multi-modal language model Chameleon~\cite{team2024chameleon} in most tasks, exhibiting undegraded linguistic capabilities.
Compared with Chameleon~\cite{team2024chameleon}, which is mixed pretrained with an extremely large scale of data, 
\NAME performs training from existing LLMs that already possess decent language capabilities, maintaining these capabilities without degradation. This result validates the efficiency of our training framework and demonstrates that with this framework, we can extend the visual generation and understanding capabilities to LLMs of any structure and size.

\begin{table*}[t!]
    \centering
    \resizebox{1.0\linewidth}{!}{
    \begin{tabular}{l l l l l | c c c c c }
    \toprule
    {\bf Type} &{\bf Method} & {\bf LLM} & {\bf Visual Token} & {\bf Res.} & {\bf VQAv2} & {\bf GQA} & {\bf TextVQA} & {\bf POPE} & {\bf MME}   \\
    \midrule
    \multirow{4}{*}{Und. Only}  &LLaVA-1.5 \cite{llava} & Vicuna-1.5-7B & Continuous & 336 & 78.5$^*$ & 62.0$^*$ & 58.2 & 85.9 & 1510.7  \\
    &VILA \cite{lin2023vila} & LLaMA-2-7B & Continuous & 336 & 79.9$^*$ & 62.3$^*$ & 64.4 & 85.5 & 1533.0  \\
    &InstructBLIP \cite{instructblip} & Vicuna-7B & Continuous & 224 & -- & 49.2 & 50.1 & -- & -- \\
    &IDEFICS-9B \cite{idefics} & LLaMA-7B & Continuous & 224 & 50.9 & 38.4 & 25.9 & -- & -- \\
    \midrule
    \multirow{11}{*}{Und. \& Gen.}  &Unified-IO 2 \cite{lu2024unified} & 6.8B from scratch & Continuous & 384 & 79.4$^*$ & -- & -- & 87.7 & --  \\
    &Emu \cite{sun2023generative} & LLaMA-13B & Continuous & 224 & 52.0 & -- & -- & -- & -- \\
    &LaVIT \cite{jin2023unified} & LLaMA-7B & Continuous & 224 & 66.0 & 46.8 & -- & -- & -- \\
    &DreamLLM \cite{dreamllm} & Vicuna-7B & Continuous & 224 & 72.9$^*$ & -- & 41.8 & -- & -- \\
    &CM3Leon-7B \cite{CM3Leon} & 7B from scratch & Discrete & 256 & 47.6 & -- & -- & -- & --  \\
    &LWM \cite{liu2024world} & LLaMA-2-7B & Discrete & 256 & 55.8 & 44.8 & 18.8 & 75.2 & -- \\
    &Show-o \cite{xie2024show} & Phi-1.5-1.3B & Discrete & 256 & 59.3$^*$ & 48.7$^*$ & -- & 73.8 & 948.4  \\
    &VILA-U~\cite{liu2024world} & LLaMA-2-7B & Discrete & 256 & 75.3$^*$ & 58.3$^*$ & 48.3 & 83.9 & 1336.2   \\
    &Chameleon~\cite{team2024chameleon} &34B from scratch &Discrete &512 &69.6 & -- & -- & -- & --   \\
    &Ours & Gemma-7B & Discrete  &512 &68.0$^*$  &56.1$^*$ &40.4 &81.1 &1107.2  \\
    &Ours$\dagger$ & Gemma-7B & Discrete  &512 &71.3$^*$ &58.4$^*$ &42.4 &81.1 &1119.3 \\
    &Ours$\ddagger$ &LLaMA-2-7B & Discrete  &256 &76.8$^*$ &61.1$^*$ &51.6 &83.2 &1448.0 \\
    \bottomrule
    \end{tabular}
    }
    \caption{Comparison with leading methods on visual language benchmarks. * indicates that images in the training split of these datasets are observed during training. 
    “Und.” and “Gen.” denote “understanding” and “generation”.
    Our performance surpasses most models that unify understanding and generation, and it is comparable with models dedicated to visual understanding in some tasks.
    $\dagger$ has a longer pre-training phase, $\ddagger$ indicates the use of a VQVAE aligned with CLIP semantics during training.
    }
    \label{tab:image_language_understanding}
\end{table*}

\subsection{Quantitative Results on Visual Understanding}
To evaluate the visual understanding capabilities, 
we use 1M LMSYS~\cite{zheng2023lmsys} as text-only instruction data, coupled with 1M text-to-Image data sampled from high-quality data, and 1.5M multi-modal instruction tuning data introduced in Minigemini~\cite{li2024mini}. This compiles a 3.5M hybrid instruction tuning data for further refining our pretrained model.
We report results on widely-adopted zero-shot image-based benchmarks, which include VQA-v2~\cite{vqa_v2}, GQA~\cite{gqa}, TextVQA~\cite{textvqa}, POPE~\cite{POPE}, and MME~\cite{mme}.

As shown in Tab.~\ref{tab:image_language_understanding}, compare with the MLLMs with discrete visual token,
\NAME outperforms models with stander VQVAE like LWM~\cite{liu2024world}, Chameleon~\cite{team2024chameleon}, and Show-o~\cite{xie2024show}.
However, the performance of MLLMs using discrete visual tokens on visual understanding tasks tends to be lower than mainstream models that employ continuous visual tokens. 
Most MLLMs with continuous visual token use CLIP features as visual input, it is a significant advantage considering that CLIP is pre-trained on a large-scale image-text pair dataset, leading to a strong alignment between its visual space features and language space features. This alignment substantially aids the LLMs, making it easier for them to understand visual content. In contrast, using image tokens derived directly from VQVAE tokenizer as input means that the corresponding embedding features in the LLM are reinitialized with out any alignment. Without extensive pre-training to align feature spaces, the visual understanding capabilities might be slightly inferior to models using CLIP as visual input. The difference in performance mainly stems from the fact that most VQVAE currently do not align image-text spaces. VILA-U~\cite{wu2024vila} has confirmed that by adding CLIP loss during the VQVAE training to align its visual space, the performance of visual understanding tasks can be improved.

In another hand, we attempt to increase the training step the pre-training phase by add one more epoch, and then we were surprised to discover that the model could achieve better performance on visual understanding tasks (marked with $\dagger$). This result indicates that mixed-modality pre-training plays a role similar to CLIP pre-training, aligning text and visual embedding spaces. More pret-raining or more suitable embedding initialization methods could further boost the performance of discrete visual tokens on visual understanding tasks.
To further explore the potential of this paradigm, we replace the VQVAE in Chameleon with UniTok~\cite{ma2025unitok} that aligns visual and language spaces during training (marked with $\ddagger$), outperforming VILA-U~\cite{wu2024vila}, which trained a CLIP-based multi-codebook VQVAE to improve understanding, and achieving results on par with LLaVA. This demonstrates the importance of visual-semantic space alignment for understanding tasks. Larger-scale pretraining or introducing semantic-aligned priors for visual tokens proves crucial for enhancing the comprehension capabilities of unified multimodal models, representing a key direction for future tokenizer improvements.


\subsection{In-Context Learning Across Modalities}



\begin{wrapfigure}{r}{7.5cm}
\vspace{-2mm}
\centering
\includegraphics[width=7.5cm]{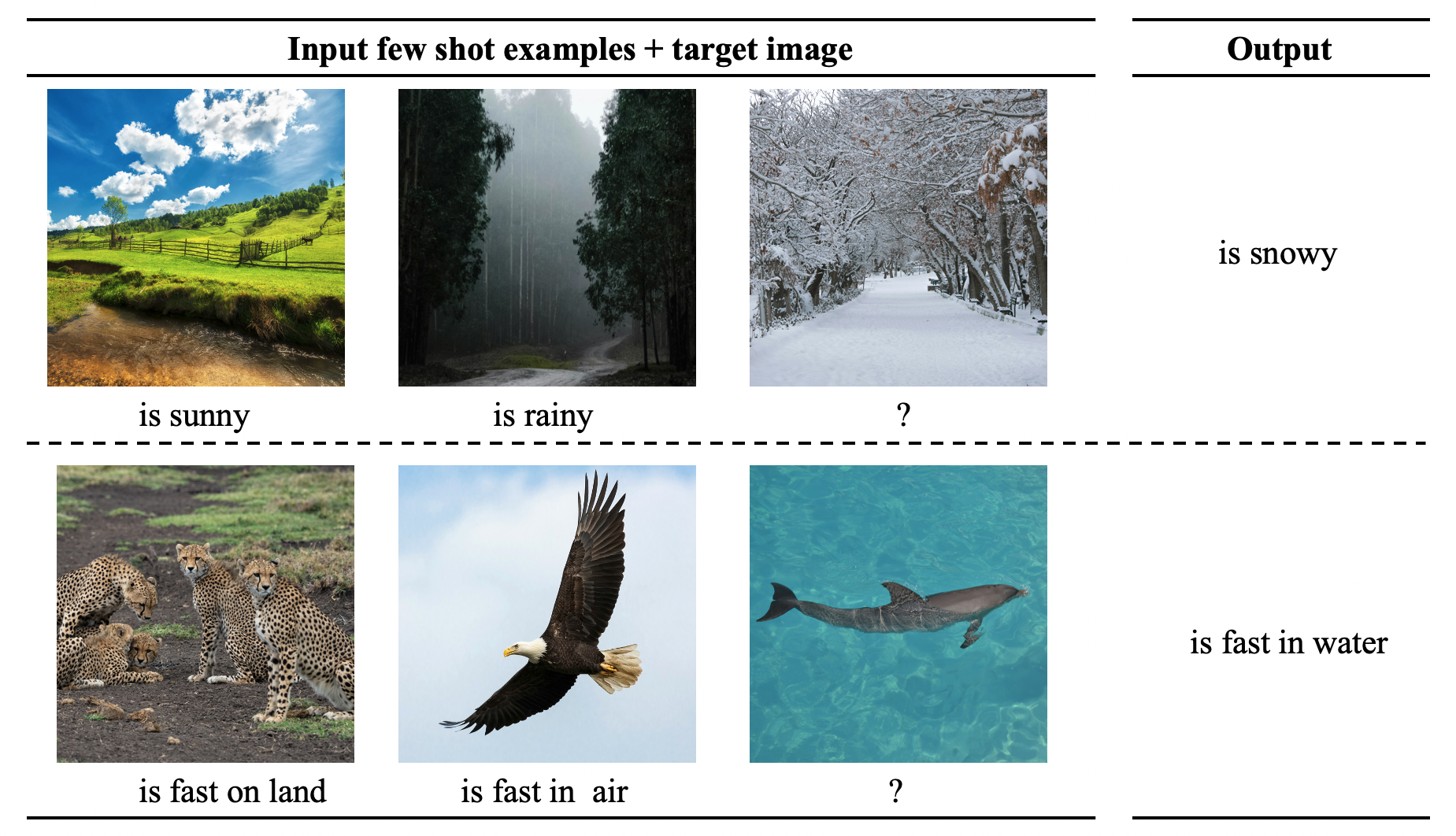} 
\caption{\NAME has good in-context learning capability. We feed two image-text pairs and a third image as the context to prompt the model.}
\label{fig:ICL}
\end{wrapfigure}


In-context learning (ICL) is a hallmark capability of LLMs that enables few-shot adaptation through task demonstrations — has revolutionized how models generalize to unseen scenarios without parameter updates. To investigate whether our unified multimodal architecture \NAME can extend this emergent capability to the vision-language space, we attempt to provide the model with several cross-modal task examples.
We find that incorporating interleaved multimodal data~\cite{mmc4} into the pretraining dataset enables the model to further learn interactions between multimodal contents, as shown in Fig.~\ref{fig:ICL}, given ``\texttt{<image1>} is sunny, \texttt{<image2>} is rainy,  \texttt{<image3>}'', 
\NAME correctly associates the third image with ``is snowy'' by recognizing the weather-related visual-textual pattern. Similarly, when shown images of a cheetah, eagle, and dolphin paired with their motion domains ``fast on land'', ``fast in air'', the model infers ``fast in water'' 
by aligning visual semantics with spatial attributes. These results demonstrate that the model not only grounds individual modalities but also discovers structured cross-modal relationships from limited context, mimicking the compositional reasoning of pure-text ICL. 
Crucially, this capability emerges without explicit multi-modal alignment supervision, suggesting that the model internalizes a unified representation space where visual and textual patterns cohere into reusable "prompts". 
Our findings position multi-modal ICL as a promising direction for few-shot adaptation in vision-language systems.

\subsection{Visual Comparative Analysis}

\begin{figure*}[tb]
\centering
\includegraphics[width=1.0 \linewidth]{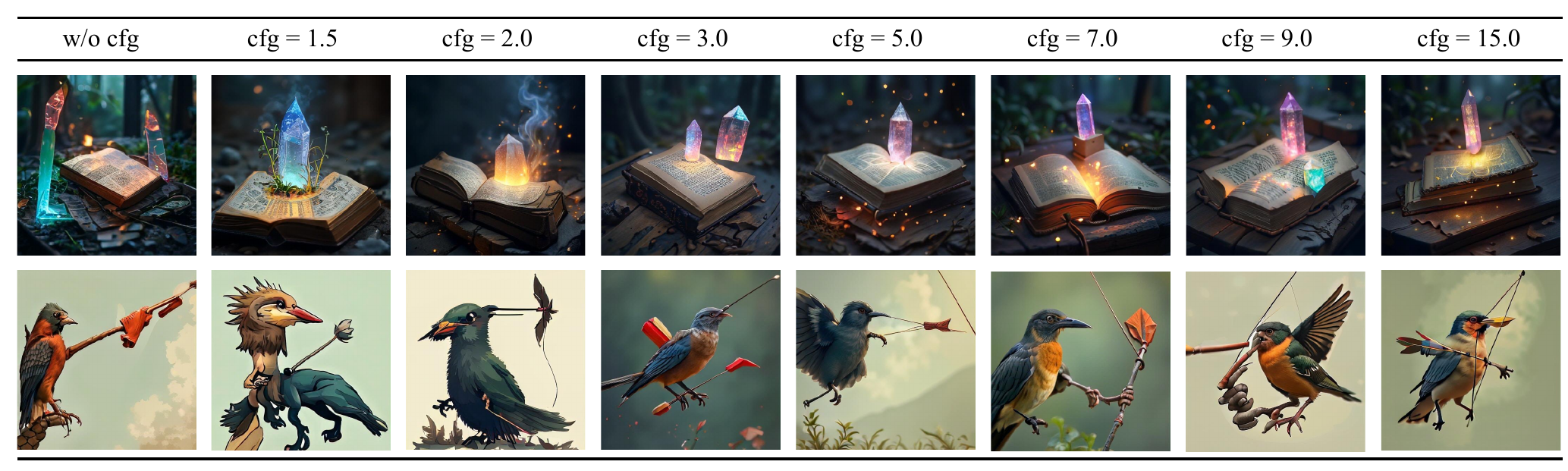}
\caption{The prompt for the first row of images is "A book with glowing runes floating beside a mystic crystal." The prompt for the second row is "A bird nocks an arrow." Higher CFG scales enhance the consistency between the model and the text prompt but degrade photorealism, while lower CFG scales improve image realism at the cost of weaker semantic alignment.}
\label{fig:cfg}
\end{figure*}

\textbf{Impact of Classifier-free Guidance.} 
Classifier-Free Guidance (CFG) scale is hyperparameter that control the trade-off between sample quality and diversity in conditional generative models. The visual variations of generated images with different CFG scales t are illustrated in Fig.~\ref{fig:cfg}
As observed, higher CFG scales lead to better alignment between the generated images and the text prompts, but cause more chaotic object structures, stronger stylization, and worse photorealism. For example, when CFG=15, the structure of the book in the image becomes disordered. Conversely, lower CFG scales result in poorer consistency between the image content and the prompt, but improve the photorealism and fine-grained texture details.

\begin{figure*}[tb]
\centering
\includegraphics[width=1.0 \linewidth]{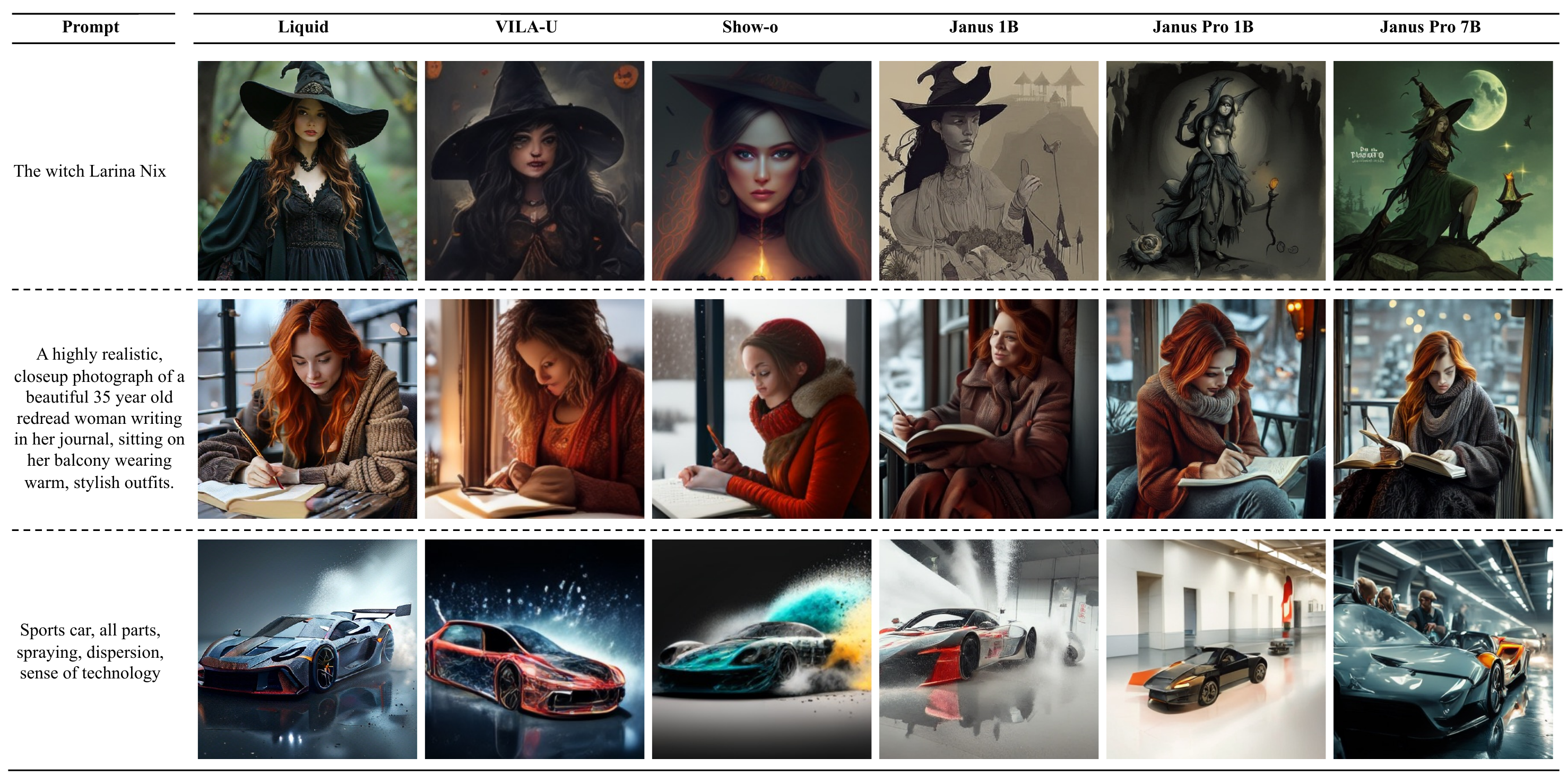}
\caption{Visual generation comparison between Liquid and other unified multim-odal models for understanding and generation}
\label{fig:model_compare}
\end{figure*}

\textbf{Comparison with Other Models.} In Fig.~\ref{fig:model_compare}, we present a comparison between Liquid and other unified multi-modal large models in terms of visual generation quality. Compared to models based on discrete multi-codebook (VILA-U), diffusion processes (Show-o), and multimodal tokenizers (Janus), Liquid demonstrates superior performance in knowledge-aware image generation (first row), scene generation accuracy and small-scale facial details (second row), as well as structural coherence of objects (third row, vehicles).
This visual comparison quantitatively validates Liquid's advancements in generating high-fidelity, semantically consistent images while maintaining strong multi-modal understanding capabilities.

\subsection{Discussion}
Liquid addresses the following limitations of previous works:

1. Previous unified multimodal models usually suffered from degraded language capabilities, limiting their broader applicability. Liquid demonstrates that a unified multimodal model can maintain on-par language performance even after continued training, preserving its potential as a versatile foundation model.  

2. No prior work has explored whether LLMs retain the power-law scaling laws observed in language tasks when extended to visual generation tasks. We prove this alignment and further show that vision can be effectively learned by LLMs as a form of language.  

3. Previous works~\cite{pan2024auto,wu2024janus} observed conflicts between visual understanding and generation tasks. We discover that the unified token space enables visual generation and comprehension tasks to mutually enhance each other, effectively removing the conflict.

\section{Related Work}
\label{sec:Related}

\textbf{Multi-modal Large Language Models.}
The rapid advancement of Large Language Models (LLMs)~\cite{PALM, gpt3, llama, touvron2023llama, team2024gemma} in recent years has inspired researchers to explore their application in visual understanding tasks. The integration of visual information with language models brings about potent multi-modal comprehension and reasoning abilities. Initial works such as LLaVA~\cite{llava} and MiniGPT4~\cite{zhu2023minigpt} propose to project features from a pre-trained visual foundation model~\cite{CLIP, li2022blip} into the feature space of LLMs, exhibiting encouraging multi-modal understanding capacities. Building upon this progress, an array of MLLMs~\cite{flamingo, li2023blip, instructblip, bai2023qwen} have been well-designed and extensively trained on comprehensive vision-language data, achieving noteworthy performance on visual understanding and reasoning tasks. The LLaVA series~\cite{llava, liu2024improved, lu2023empirical, chen2024allava, chen2023sharegpt4v, liu2024llavanext, li2024mini} employ image-text pair data to train a projector, projecting the image-feature from CLIP to align the language spaces within the input space of LLMs. They further enhance visual understanding and reasoning abilities by training the entire pipeline via a curated multi-modal instruction tuning dataset. Despite their robust multi-modal understanding capabilities, existing models are primarily focused on visual understanding, falling short on generating visual outputs that extend beyond text.

\textbf{Vision Generation.}
In the past few years, the realm of visual generation has been primarily dominated by diffusion models~\cite{sdxl, ldm, midjourney, dalle-3, DALLE2}, which progressively generate high-quality, high-resolution images via a diffusion process over a continuous latent space. Several efforts~\cite{ge2023planting, ge2024seed, jin2023unified, sun2023generative, sun2024generative,loong} have attempted to extend LLMs with pretrained diffusion models to integrate image generation capabilities. These studies employ diffusion models as a tool where the diffusion models generate images conditioned on the features output by the LLMs. In this combination, LLMs merely contribute the semantic feature output and lack the direct ability to generate visual content. Moreover, the upper limit of visual generation capacity is dictated by the pre-trained diffusion model, leaving the inherent potential of LLMs in visual generation under-explored. 

An alternative viable approach involves using autoregressive models to generate images by predicting the next token in a sequence, as exemplified by models like DALL-E~\cite{DALLE}, CogView~\cite{ding2021cogview}, Parti~\cite{yu2022scaling} and LlamaGen~\cite{sun2024autoregressive}. Visual AutoRegressive modeling (VAR)\cite{VAR} redefined auto-regressive learning on images as coarse-to-fine “next-scale prediction”. It demonstrates superior generalization and scaling capabilities compared to diffusion transformers while requiring fewer steps.These models typically employ VQVAE~\cite{esser2021taming} to tokenize images into a set of discrete codes, subsequently training a decoder-only transformer to predict image codes which are then detokenized back to images. These approaches showcase the potential of decoder-only LLMs in directly conducting image generation. However, they often fail to match the performance of diffusion models and do not explore the possibility of unified output between visual and linguistic modalities. In this work, our objective is to enable LLMs to generate visual content via next-token prediction without altering their structure or capabilities, and explore the characteristics that emerge from the combination of these two tasks within LLMs.

\textbf{Unified Multimodal Understanding and Generation}
Several early efforts have explored how to construct a unified multi-modal large model for visual generation and understanding based on LLMs. The central challenge lies in tokenizing images into sequence inputs for the LLMs and detokenizing the sequential output of the LLMs back into images, the choice of image tokenizer. Some methods~\cite{ge2023planting, ge2024seed, sun2023generative, sun2024generative} use vision encoders based on ViT like CLIP to encode images into continuous feature maps. The continuous visual space from CLIP can retain more visual information and have a pre-trained, aligned space with language feature. However, the continuous feature often necessitates an additional diffusion module for image detokenization. 
Other works ~\cite{liu2024world, team2024chameleon, wu2024vila} employ VQVAE to encode images into discrete tokens and train LLMs to predict them. Still other works~\cite{lu2024unified, xie2024show, wu2024janus, qu2024tokenflow} use both ViT and VQVAE as tokenizers to garner their benefits. Discrete image features can share the same embedding space with text input, permitting joint reasoning over both modalities within a unified architecture without the requirement for modality-specific components. It is beneficial for model scale-up. Consequently, in our work, we choose VQVAE as the sole image tokenizer. 
Our work is most similar to LWM~\cite{liu2024world} and Chameleon~\cite{team2024chameleon}. However, they display inferior image understanding and generation capabilities, and need extensive large scale multi-modal pre-training, which is a significant burden. In contrast, we propose to start from any existing LLMs and enhance their visual understanding and generation abilities by continuing training with a small amount of high-quality data, without altering any model structures.

\section{Conclusion}
In this paper, we present \NAME, an efficient framework enabling language models to acquire image generation and understanding capabilities without modifying the original structure. Unlike traditional multi-modal models employing extra visual models, \NAME directly tokenize images into discrete tokens that share the same embedding space with text tokens. This leads to a total unification of images and text within the models, which stokes the potential of multi-modal learning. Utilizing various existing LLMs provides a unique advantage to \NAME, enabling it to scale up easily and display similar scaling behavior to LLMs. 

Leveraging this convenience, we conducted extensive scaling experiments on models ranging from 0.5B to 32B across different model families. We identified some key characteristics of multimodal models under this unified token space. 1) Firstly, by directly training LLMs on visual generation tasks, they can retain foundational language capabilities while achieving results comparable to some mainstream diffusion models. 2) Secondly, this unification of multimodal tasks tends to impair both visual generation and language tasks; however, this impairment gradually diminishes as the model size increases. 3) Lastly, we found that when visual and language tokens are represented uniformly, visual understanding and generation tasks can mutually enhance each other. This reciprocity encourages the vast potential of large-scale pretraining under this paradigm.

\clearpage
{
\small

\bibliographystyle{cite}
\bibliography{cite}

\begin{thebibliography}{10}\itemsep=-1pt

\bibitem{flamingo}
J.-B. Alayrac, J.~Donahue, P.~Luc, A.~Miech, I.~Barr, Y.~Hasson, K.~Lenc, A.~Mensch, K.~Millican, M.~Reynolds, et~al.
\newblock Flamingo: a visual language model for few-shot learning.
\newblock {\em Advances in Neural Information Processing Systems}, 35:23716--23736, 2022.

\bibitem{bai2023qwen}
J.~Bai, S.~Bai, S.~Yang, S.~Wang, S.~Tan, P.~Wang, J.~Lin, C.~Zhou, and J.~Zhou.
\newblock Qwen-vl: A frontier large vision-language model with versatile abilities.
\newblock {\em arXiv preprint arXiv:2308.12966}, 2023.

\bibitem{bisk2020piqa}
Y.~Bisk, R.~Zellers, J.~Gao, Y.~Choi, et~al.
\newblock Piqa: Reasoning about physical commonsense in natural language.
\newblock In {\em Proceedings of the AAAI conference on artificial intelligence}, pages 7432--7439, 2020.

\bibitem{gpt3}
T.~Brown, B.~Mann, N.~Ryder, M.~Subbiah, J.~D. Kaplan, P.~Dhariwal, A.~Neelakantan, P.~Shyam, G.~Sastry, A.~Askell, et~al.
\newblock Language models are few-shot learners.
\newblock {\em Advances in neural information processing systems}, 33:1877--1901, 2020.

\bibitem{chen2024allava}
G.~H. Chen, S.~Chen, R.~Zhang, J.~Chen, X.~Wu, Z.~Zhang, Z.~Chen, J.~Li, X.~Wan, and B.~Wang.
\newblock Allava: Harnessing gpt4v-synthesized data for a lite vision-language model.
\newblock {\em arXiv preprint arXiv:2402.11684}, 2024.

\bibitem{chen2023pixart}
J.~Chen, J.~Yu, C.~Ge, L.~Yao, E.~Xie, Y.~Wu, Z.~Wang, J.~Kwok, P.~Luo, H.~Lu, et~al.
\newblock Pixart-$alpha$: Fast training of diffusion transformer for photorealistic text-to-image synthesis.
\newblock {\em arXiv preprint arXiv:2310.00426}, 2023.

\bibitem{chen2023sharegpt4v}
L.~Chen, J.~Li, X.~Dong, P.~Zhang, C.~He, J.~Wang, F.~Zhao, and D.~Lin.
\newblock Sharegpt4v: Improving large multi-modal models with better captions.
\newblock {\em arXiv preprint arXiv:2311.12793}, 2023.

\bibitem{PALM}
A.~Chowdhery, S.~Narang, J.~Devlin, M.~Bosma, G.~Mishra, A.~Roberts, P.~Barham, H.~W. Chung, C.~Sutton, S.~Gehrmann, et~al.
\newblock Palm: Scaling language modeling with pathways.
\newblock {\em arXiv preprint arXiv:2204.02311}, 2022.

\bibitem{clark2019boolq}
C.~Clark, K.~Lee, M.-W. Chang, T.~Kwiatkowski, M.~Collins, and K.~Toutanova.
\newblock Boolq: Exploring the surprising difficulty of natural yes/no questions.
\newblock {\em arXiv preprint arXiv:1905.10044}, 2019.

\bibitem{clark2018think}
P.~Clark, I.~Cowhey, O.~Etzioni, T.~Khot, A.~Sabharwal, C.~Schoenick, and O.~Tafjord.
\newblock Think you have solved question answering? try arc, the ai2 reasoning challenge.
\newblock {\em arXiv preprint arXiv:1803.05457}, 2018.

\bibitem{instructblip}
W.~Dai, J.~Li, D.~Li, A.~M.~H. Tiong, J.~Zhao, W.~Wang, B.~Li, P.~Fung, and S.~Hoi.
\newblock Instructblip: Towards general-purpose vision-language models with instruction tuning.
\newblock {\em arXiv}, 2023.

\bibitem{ding2021cogview}
M.~Ding, Z.~Yang, W.~Hong, W.~Zheng, C.~Zhou, D.~Yin, J.~Lin, X.~Zou, Z.~Shao, H.~Yang, et~al.
\newblock Cogview: Mastering text-to-image generation via transformers.
\newblock {\em Advances in neural information processing systems}, 34:19822--19835, 2021.

\bibitem{dreamllm}
R.~Dong, C.~Han, Y.~Peng, Z.~Qi, Z.~Ge, J.~Yang, L.~Zhao, J.~Sun, H.~Zhou, H.~Wei, et~al.
\newblock Dreamllm: Synergistic multimodal comprehension and creation.
\newblock {\em arXiv preprint arXiv:2309.11499}, 2023.

\bibitem{dubey2024llama}
A.~Dubey, A.~Jauhri, A.~Pandey, A.~Kadian, A.~Al-Dahle, A.~Letman, A.~Mathur, A.~Schelten, A.~Yang, A.~Fan, et~al.
\newblock The llama 3 herd of models.
\newblock {\em arXiv preprint arXiv:2407.21783}, 2024.

\bibitem{esser2021taming}
P.~Esser, R.~Rombach, and B.~Ommer.
\newblock Taming transformers for high-resolution image synthesis.
\newblock In {\em Proceedings of the IEEE/CVF conference on computer vision and pattern recognition}, pages 12873--12883, 2021.

\bibitem{mme}
C.~Fu, P.~Chen, Y.~Shen, Y.~Qin, M.~Zhang, X.~Lin, J.~Yang, X.~Zheng, K.~Li, X.~Sun, Y.~Wu, and R.~Ji.
\newblock Mme: A comprehensive evaluation benchmark for multimodal large language models, 2024.

\bibitem{ge2023planting}
Y.~Ge, Y.~Ge, Z.~Zeng, X.~Wang, and Y.~Shan.
\newblock Planting a seed of vision in large language model.
\newblock {\em arXiv preprint arXiv:2307.08041}, 2023.

\bibitem{ge2024seed}
Y.~Ge, S.~Zhao, J.~Zhu, Y.~Ge, K.~Yi, L.~Song, C.~Li, X.~Ding, and Y.~Shan.
\newblock Seed-x: Multimodal models with unified multi-granularity comprehension and generation.
\newblock {\em arXiv preprint arXiv:2404.14396}, 2024.

\bibitem{vqa_v2}
Y.~Goyal, T.~Khot, D.~Summers{-}Stay, D.~Batra, and D.~Parikh.
\newblock Making the {V} in {VQA} matter: Elevating the role of image understanding in {V}isual {Q}uestion {A}nswering.
\newblock In {\em Conference on Computer Vision and Pattern Recognition (CVPR)}, 2017.

\bibitem{Hendrycks2020MeasuringMM}
D.~Hendrycks, C.~Burns, S.~Basart, A.~Zou, M.~Mazeika, D.~X. Song, and J.~Steinhardt.
\newblock Measuring massive multitask language understanding.
\newblock {\em arXiv preprint arXiv:2009.03300}, 2020.

\bibitem{fid}
M.~Heusel, H.~Ramsauer, T.~Unterthiner, B.~Nessler, and S.~Hochreiter.
\newblock Gans trained by a two time-scale update rule converge to a local nash equilibrium.
\newblock {\em Advances in neural information processing systems}, 30, 2017.

\bibitem{gqa}
D.~A. Hudson and C.~D. Manning.
\newblock Gqa: A new dataset for real-world visual reasoning and compositional question answering.
\newblock In {\em Proceedings of the IEEE/CVF conference on computer vision and pattern recognition}, pages 6700--6709, 2019.

\bibitem{hui2024qwen2}
B.~Hui, J.~Yang, Z.~Cui, J.~Yang, D.~Liu, L.~Zhang, T.~Liu, J.~Zhang, B.~Yu, K.~Lu, et~al.
\newblock Qwen2. 5-coder technical report.
\newblock {\em arXiv preprint arXiv:2409.12186}, 2024.

\bibitem{jin2023unified}
Y.~Jin, K.~Xu, L.~Chen, C.~Liao, J.~Tan, B.~Chen, C.~Lei, A.~Liu, C.~Song, X.~Lei, et~al.
\newblock Unified language-vision pretraining with dynamic discrete visual tokenization.
\newblock {\em arXiv preprint arXiv:2309.04669}, 2023.

\bibitem{idefics}
H.~Lauren{\c{c}}on, D.~van Strien, S.~Bekman, L.~Tronchon, L.~Saulnier, T.~Wang, S.~Karamcheti, A.~Singh, G.~Pistilli, Y.~Jernite, et~al.
\newblock Introducing idefics: An open reproduction of state-of-the-art visual language model, 2023.
\newblock {\em URL https://huggingface. co/blog/idefics. Accessed}, pages 09--18, 2023.

\bibitem{playgroundv2.5}
D.~Li, A.~Kamko, E.~Akhgari, A.~Sabet, L.~Xu, and S.~Doshi.
\newblock Playground v2. 5: Three insights towards enhancing aesthetic quality in text-to-image generation.
\newblock {\em arXiv preprint arXiv:2402.17245}, 2024.

\bibitem{li2024synergen}
H.~Li, C.~Tian, J.~Shao, X.~Zhu, Z.~Wang, J.~Zhu, W.~Dou, X.~Wang, H.~Li, L.~Lu, et~al.
\newblock Synergen-vl: Towards synergistic image understanding and generation with vision experts and token folding.
\newblock {\em arXiv preprint arXiv:2412.09604}, 2024.

\bibitem{li2024datacomplm}
J.~Li, A.~Fang, G.~Smyrnis, M.~Ivgi, M.~Jordan, S.~Gadre, H.~Bansal, E.~Guha, S.~Keh, K.~Arora, S.~Garg, R.~Xin, N.~Muennighoff, R.~Heckel, J.~Mercat, M.~Chen, S.~Gururangan, M.~Wortsman, A.~Albalak, Y.~Bitton, M.~Nezhurina, A.~Abbas, C.-Y. Hsieh, D.~Ghosh, J.~Gardner, M.~Kilian, H.~Zhang, R.~Shao, S.~Pratt, S.~Sanyal, G.~Ilharco, G.~Daras, K.~Marathe, A.~Gokaslan, J.~Zhang, K.~Chandu, T.~Nguyen, I.~Vasiljevic, S.~Kakade, S.~Song, S.~Sanghavi, F.~Faghri, S.~Oh, L.~Zettlemoyer, K.~Lo, A.~El-Nouby, H.~Pouransari, A.~Toshev, S.~Wang, D.~Groeneveld, L.~Soldaini, P.~W. Koh, J.~Jitsev, T.~Kollar, A.~G. Dimakis, Y.~Carmon, A.~Dave, L.~Schmidt, and V.~Shankar.
\newblock Datacomp-lm: In search of the next generation of training sets for language models.
\newblock {\em arXiv preprint arXiv:2406.11794}, 2024.

\bibitem{li2023blip}
J.~Li, D.~Li, S.~Savarese, and S.~Hoi.
\newblock Blip-2: Bootstrapping language-image pre-training with frozen image encoders and large language models.
\newblock In {\em International conference on machine learning}, pages 19730--19742. PMLR, 2023.

\bibitem{li2022blip}
J.~Li, D.~Li, C.~Xiong, and S.~Hoi.
\newblock Blip: Bootstrapping language-image pre-training for unified vision-language understanding and generation.
\newblock In {\em International conference on machine learning}, pages 12888--12900. PMLR, 2022.

\bibitem{li2023starcoder}
R.~Li, L.~B. Allal, Y.~Zi, N.~Muennighoff, D.~Kocetkov, C.~Mou, M.~Marone, C.~Akiki, J.~Li, J.~Chim, Q.~Liu, E.~Zheltonozhskii, T.~Y. Zhuo, T.~Wang, O.~Dehaene, M.~Davaadorj, J.~Lamy-Poirier, J.~Monteiro, O.~Shliazhko, N.~Gontier, N.~Meade, A.~Zebaze, M.-H. Yee, L.~K. Umapathi, J.~Zhu, B.~Lipkin, M.~Oblokulov, Z.~Wang, R.~Murthy, J.~Stillerman, S.~S. Patel, D.~Abulkhanov, M.~Zocca, M.~Dey, Z.~Zhang, N.~Fahmy, U.~Bhattacharyya, W.~Yu, S.~Singh, S.~Luccioni, P.~Villegas, M.~Kunakov, F.~Zhdanov, M.~Romero, T.~Lee, N.~Timor, J.~Ding, C.~Schlesinger, H.~Schoelkopf, J.~Ebert, T.~Dao, M.~Mishra, A.~Gu, J.~Robinson, C.~J. Anderson, B.~Dolan-Gavitt, D.~Contractor, S.~Reddy, D.~Fried, D.~Bahdanau, Y.~Jernite, C.~M. Ferrandis, S.~Hughes, T.~Wolf, A.~Guha, L.~von Werra, and H.~de~Vries.
\newblock Starcoder: may the source be with you!
\newblock {\em arXiv preprint arXiv:2305.06161}, 2023.

\bibitem{POPE}
Y.~Li, Y.~Du, K.~Zhou, J.~Wang, W.~X. Zhao, and J.-R. Wen.
\newblock Evaluating object hallucination in large vision-language models.
\newblock {\em arXiv preprint arXiv:2305.10355}, 2023.

\bibitem{li2024mini}
Y.~Li, Y.~Zhang, C.~Wang, Z.~Zhong, Y.~Chen, R.~Chu, S.~Liu, and J.~Jia.
\newblock Mini-gemini: Mining the potential of multi-modality vision language models.
\newblock {\em arXiv preprint arXiv:2403.18814}, 2024.

\bibitem{lin2023vila}
J.~Lin, H.~Yin, W.~Ping, Y.~Lu, P.~Molchanov, A.~Tao, H.~Mao, J.~Kautz, M.~Shoeybi, and S.~Han.
\newblock Vila: On pre-training for visual language models, 2023.

\bibitem{GenAI-Bench}
Z.~Lin, D.~Pathak, B.~Li, J.~Li, X.~Xia, G.~Neubig, P.~Zhang, and D.~Ramanan.
\newblock Evaluating text-to-visual generation with image-to-text generation.
\newblock {\em arXiv preprint arXiv:2404.01291}, 2024.

\bibitem{dalle-3}
Z.~Lin, D.~Pathak, B.~Li, J.~Li, X.~Xia, G.~Neubig, P.~Zhang, and D.~Ramanan.
\newblock Evaluating text-to-visual generation with image-to-text generation.
\newblock {\em arXiv preprint arXiv:2404.01291}, 2024.

\bibitem{liu2024improved}
H.~Liu, C.~Li, Y.~Li, and Y.~J. Lee.
\newblock Improved baselines with visual instruction tuning.
\newblock In {\em Proceedings of the IEEE/CVF Conference on Computer Vision and Pattern Recognition}, pages 26296--26306, 2024.

\bibitem{liu2024llavanext}
H.~Liu, C.~Li, Y.~Li, B.~Li, Y.~Zhang, S.~Shen, and Y.~J. Lee.
\newblock Llava-next: Improved reasoning, ocr, and world knowledge, January 2024.

\bibitem{llava}
H.~Liu, C.~Li, Q.~Wu, and Y.~J. Lee.
\newblock Visual instruction tuning.
\newblock {\em Advances in neural information processing systems}, 36, 2024.

\bibitem{liu2024world}
H.~Liu, W.~Yan, M.~Zaharia, and P.~Abbeel.
\newblock World model on million-length video and language with ringattention.
\newblock {\em arXiv preprint arXiv:2402.08268}, 2024.

\bibitem{lu2024unified}
J.~Lu, C.~Clark, S.~Lee, Z.~Zhang, S.~Khosla, R.~Marten, D.~Hoiem, and A.~Kembhavi.
\newblock Unified-io 2: Scaling autoregressive multimodal models with vision language audio and action.
\newblock In {\em Proceedings of the IEEE/CVF Conference on Computer Vision and Pattern Recognition}, pages 26439--26455, 2024.

\bibitem{lu2023empirical}
Y.~Lu, C.~Li, H.~Liu, J.~Yang, J.~Gao, and Y.~Shen.
\newblock An empirical study of scaling instruct-tuned large multimodal models.
\newblock {\em arXiv preprint arXiv:2309.09958}, 2023.

\bibitem{ma2025unitok}
C.~Ma, Y.~Jiang, J.~Wu, J.~Yang, X.~Yu, Z.~Yuan, B.~Peng, and X.~Qi.
\newblock Unitok: A unified tokenizer for visual generation and understanding.
\newblock {\em arXiv preprint arXiv:2502.20321}, 2025.

\bibitem{mihaylov2018can}
T.~Mihaylov, P.~Clark, T.~Khot, and A.~Sabharwal.
\newblock Can a suit of armor conduct electricity? a new dataset for open book question answering.
\newblock {\em arXiv preprint arXiv:1809.02789}, 2018.

\bibitem{niu2025wise}
Y.~Niu, M.~Ning, M.~Zheng, B.~Lin, P.~Jin, J.~Liao, K.~Ning, B.~Zhu, and L.~Yuan.
\newblock Wise: A world knowledge-informed semantic evaluation for text-to-image generation.
\newblock {\em arXiv preprint arXiv:2503.07265}, 2025.

\bibitem{pan2023journeydb}
J.~Pan, K.~Sun, Y.~Ge, H.~Li, H.~Duan, X.~Wu, R.~Zhang, A.~Zhou, Z.~Qin, Y.~Wang, J.~Dai, Y.~Qiao, and H.~Li.
\newblock Journeydb: A benchmark for generative image understanding, 2023.

\bibitem{pan2024auto}
K.~Pan, S.~Tang, J.~Li, Z.~Fan, W.~Chow, S.~Yan, T.-S. Chua, Y.~Zhuang, and H.~Zhang.
\newblock Auto-encoding morph-tokens for multimodal llm.
\newblock {\em arXiv preprint arXiv:2405.01926}, 2024.

\bibitem{sdxl}
D.~Podell, Z.~English, K.~Lacey, A.~Blattmann, T.~Dockhorn, J.~M{\"u}ller, J.~Penna, and R.~Rombach.
\newblock Sdxl: Improving latent diffusion models for high-resolution image synthesis.
\newblock {\em arXiv preprint arXiv:2307.01952}, 2023.

\bibitem{qu2024tokenflow}
L.~Qu, H.~Zhang, Y.~Liu, X.~Wang, Y.~Jiang, Y.~Gao, H.~Ye, D.~K. Du, Z.~Yuan, and X.~Wu.
\newblock Tokenflow: Unified image tokenizer for multimodal understanding and generation.
\newblock {\em arXiv preprint arXiv:2412.03069}, 2024.

\bibitem{CLIP}
A.~Radford, J.~W. Kim, C.~Hallacy, A.~Ramesh, G.~Goh, S.~Agarwal, G.~Sastry, A.~Askell, P.~Mishkin, J.~Clark, et~al.
\newblock Learning transferable visual models from natural language supervision.
\newblock In {\em International conference on machine learning}, pages 8748--8763. PMLR, 2021.

\bibitem{midjourney}
A.~M. Radhakrishnan.
\newblock Is midjourney-ai the new anti-hero of architectural imagery \& creativity?
\newblock {\em GSJ}, 11(1):94--104, 2023.

\bibitem{T5}
C.~Raffel, N.~Shazeer, A.~Roberts, K.~Lee, S.~Narang, M.~Matena, Y.~Zhou, W.~Li, and P.~J. Liu.
\newblock Exploring the limits of transfer learning with a unified text-to-text transformer.
\newblock {\em The Journal of Machine Learning Research}, 21(1):5485--5551, 2020.

\bibitem{DALLE2}
A.~Ramesh, P.~Dhariwal, A.~Nichol, C.~Chu, and M.~Chen.
\newblock Hierarchical text-conditional image generation with clip latents.
\newblock {\em arXiv preprint arXiv:2204.06125}, 1(2):3, 2022.

\bibitem{DALLE}
A.~Ramesh, M.~Pavlov, G.~Goh, S.~Gray, C.~Voss, A.~Radford, M.~Chen, and I.~Sutskever.
\newblock Zero-shot text-to-image generation.
\newblock In {\em International Conference on Machine Learning}, pages 8821--8831. PMLR, 2021.

\bibitem{ldm}
R.~Rombach, A.~Blattmann, D.~Lorenz, P.~Esser, and B.~Ommer.
\newblock High-resolution image synthesis with latent diffusion models.
\newblock In {\em Proceedings of the IEEE/CVF conference on computer vision and pattern recognition}, pages 10684--10695, 2022.

\bibitem{sakaguchi2021winogrande}
K.~Sakaguchi, R.~L. Bras, C.~Bhagavatula, and Y.~Choi.
\newblock Winogrande: An adversarial winograd schema challenge at scale.
\newblock {\em Communications of the ACM}, 64(9):99--106, 2021.

\bibitem{sap2019socialiqa}
M.~Sap, H.~Rashkin, D.~Chen, R.~LeBras, and Y.~Choi.
\newblock Socialiqa: Commonsense reasoning about social interactions.
\newblock {\em arXiv preprint arXiv:1904.09728}, 2019.

\bibitem{bpe}
R.~Sennrich, B.~Haddow, and A.~Birch.
\newblock Neural machine translation of rare words with subword units.
\newblock In K.~Erk and N.~A. Smith, editors, {\em Proceedings of the 54th Annual Meeting of the Association for Computational Linguistics (Volume 1: Long Papers)}, pages 1715--1725, Berlin, Germany, Aug. 2016. Association for Computational Linguistics.

\bibitem{textvqa}
A.~Singh, V.~Natarajan, M.~Shah, Y.~Jiang, X.~Chen, D.~Batra, D.~Parikh, and M.~Rohrbach.
\newblock Towards vqa models that can read.
\newblock In {\em Proceedings of the IEEE/CVF conference on computer vision and pattern recognition}, pages 8317--8326, 2019.

\bibitem{cerebras2023slimpajama}
D.~Soboleva, F.~Al-Khateeb, R.~Myers, J.~R. Steeves, J.~Hestness, and N.~Dey.
\newblock {SlimPajama: A 627B token cleaned and deduplicated version of RedPajama}.
\newblock \url{https://www.cerebras.net/blog/slimpajama-a-627b-token-cleaned-and-deduplicated-version-of-redpajama}, 2023.

\bibitem{sun2024autoregressive}
P.~Sun, Y.~Jiang, S.~Chen, S.~Zhang, B.~Peng, P.~Luo, and Z.~Yuan.
\newblock Autoregressive model beats diffusion: Llama for scalable image generation.
\newblock {\em arXiv preprint arXiv:2406.06525}, 2024.

\bibitem{sun2024generative}
Q.~Sun, Y.~Cui, X.~Zhang, F.~Zhang, Q.~Yu, Y.~Wang, Y.~Rao, J.~Liu, T.~Huang, and X.~Wang.
\newblock Generative multimodal models are in-context learners.
\newblock In {\em Proceedings of the IEEE/CVF Conference on Computer Vision and Pattern Recognition}, pages 14398--14409, 2024.

\bibitem{sun2023generative}
Q.~Sun, Q.~Yu, Y.~Cui, F.~Zhang, X.~Zhang, Y.~Wang, H.~Gao, J.~Liu, T.~Huang, and X.~Wang.
\newblock Generative pretraining in multimodality.
\newblock {\em arXiv preprint arXiv:2307.05222}, 2023.

\bibitem{team2024chameleon}
C.~Team.
\newblock Chameleon: Mixed-modal early-fusion foundation models.
\newblock {\em arXiv preprint arXiv:2405.09818}, 2024.

\bibitem{team2024gemma}
G.~Team, T.~Mesnard, C.~Hardin, R.~Dadashi, S.~Bhupatiraju, S.~Pathak, L.~Sifre, M.~Rivi{\`e}re, M.~S. Kale, J.~Love, et~al.
\newblock Gemma: Open models based on gemini research and technology.
\newblock {\em arXiv preprint arXiv:2403.08295}, 2024.

\bibitem{team2024gemma2}
G.~Team, M.~Riviere, S.~Pathak, P.~G. Sessa, C.~Hardin, S.~Bhupatiraju, L.~Hussenot, T.~Mesnard, B.~Shahriari, A.~Ram{\'e}, et~al.
\newblock Gemma 2: Improving open language models at a practical size.
\newblock {\em arXiv preprint arXiv:2408.00118}, 2024.

\bibitem{VAR}
K.~Tian, Y.~Jiang, Z.~Yuan, B.~Peng, and L.~Wang.
\newblock Visual autoregressive modeling: Scalable image generation via next-scale prediction.
\newblock {\em arXiv preprint arXiv:2404.02905}, 2024.

\bibitem{tong2024metamorph}
S.~Tong, D.~Fan, J.~Zhu, Y.~Xiong, X.~Chen, K.~Sinha, M.~Rabbat, Y.~LeCun, S.~Xie, and Z.~Liu.
\newblock Metamorph: Multimodal understanding and generation via instruction tuning.
\newblock {\em arXiv preprint arXiv:2412.14164}, 2024.

\bibitem{llama}
H.~Touvron, T.~Lavril, G.~Izacard, X.~Martinet, M.-A. Lachaux, T.~Lacroix, B.~Rozi{\`e}re, N.~Goyal, E.~Hambro, F.~Azhar, et~al.
\newblock Llama: Open and efficient foundation language models.
\newblock {\em arXiv preprint arXiv:2302.13971}, 2023.

\bibitem{touvron2023llama}
H.~Touvron, L.~Martin, K.~Stone, P.~Albert, A.~Almahairi, Y.~Babaei, N.~Bashlykov, S.~Batra, P.~Bhargava, S.~Bhosale, et~al.
\newblock Llama 2: Open foundation and fine-tuned chat models.
\newblock {\em arXiv preprint arXiv:2307.09288}, 2023.

\bibitem{van2017neural}
A.~Van Den~Oord, O.~Vinyals, et~al.
\newblock Neural discrete representation learning.
\newblock {\em Advances in neural information processing systems}, 30, 2017.

\bibitem{wang2024emu3}
X.~Wang, X.~Zhang, Z.~Luo, Q.~Sun, Y.~Cui, J.~Wang, F.~Zhang, Y.~Wang, Z.~Li, Q.~Yu, et~al.
\newblock Emu3: Next-token prediction is all you need.
\newblock {\em arXiv preprint arXiv:2409.18869}, 2024.

\bibitem{loong}
Y.~Wang, T.~Xiong, D.~Zhou, Z.~Lin, Y.~Zhao, B.~Kang, J.~Feng, and X.~Liu.
\newblock Loong: Generating minute-level long videos with autoregressive language models.
\newblock {\em arXiv preprint arXiv:2410.02757}, 2024.

\bibitem{wu2024janus}
C.~Wu, X.~Chen, Z.~Wu, Y.~Ma, X.~Liu, Z.~Pan, W.~Liu, Z.~Xie, X.~Yu, C.~Ruan, et~al.
\newblock Janus: Decoupling visual encoding for unified multimodal understanding and generation.
\newblock {\em arXiv preprint arXiv:2410.13848}, 2024.

\bibitem{wu2024vila}
Y.~Wu, Z.~Zhang, J.~Chen, H.~Tang, D.~Li, Y.~Fang, L.~Zhu, E.~Xie, H.~Yin, L.~Yi, et~al.
\newblock Vila-u: a unified foundation model integrating visual understanding and generation.
\newblock {\em arXiv preprint arXiv:2409.04429}, 2024.

\bibitem{xie2024show}
J.~Xie, W.~Mao, Z.~Bai, D.~J. Zhang, W.~Wang, K.~Q. Lin, Y.~Gu, Z.~Chen, Z.~Yang, and M.~Z. Shou.
\newblock Show-o: One single transformer to unify multimodal understanding and generation.
\newblock {\em arXiv preprint arXiv:2408.12528}, 2024.

\bibitem{yang2023baichuan}
A.~Yang, B.~Xiao, B.~Wang, B.~Zhang, C.~Bian, C.~Yin, C.~Lv, D.~Pan, D.~Wang, D.~Yan, et~al.
\newblock Baichuan 2: Open large-scale language models.
\newblock {\em arXiv preprint arXiv:2309.10305}, 2023.

\bibitem{yu2022scaling}
J.~Yu, Y.~Xu, J.~Y. Koh, T.~Luong, G.~Baid, Z.~Wang, V.~Vasudevan, A.~Ku, Y.~Yang, B.~K. Ayan, et~al.
\newblock Scaling autoregressive models for content-rich text-to-image generation.
\newblock {\em arXiv preprint arXiv:2206.10789}, 2(3):5, 2022.

\bibitem{CM3Leon}
L.~Yu, B.~Shi, R.~Pasunuru, B.~Muller, O.~Golovneva, T.~Wang, A.~Babu, B.~Tang, B.~Karrer, S.~Sheynin, et~al.
\newblock Scaling autoregressive multi-modal models: Pretraining and instruction tuning.
\newblock {\em arXiv preprint arXiv:2309.02591}, 2(3), 2023.

\bibitem{zellers2019hellaswag}
R.~Zellers, A.~Holtzman, Y.~Bisk, A.~Farhadi, and Y.~Choi.
\newblock Hellaswag: Can a machine really finish your sentence?
\newblock {\em arXiv preprint arXiv:1905.07830}, 2019.

\bibitem{zheng2023lmsys}
L.~Zheng, W.-L. Chiang, Y.~Sheng, T.~Li, S.~Zhuang, Z.~Wu, Y.~Zhuang, Z.~Li, Z.~Lin, E.~P. Xing, et~al.
\newblock Lmsys-chat-1m: A large-scale real-world llm conversation dataset.
\newblock {\em arXiv preprint arXiv:2309.11998}, 2023.

\bibitem{transfusion}
C.~Zhou, L.~Yu, A.~Babu, K.~Tirumala, M.~Yasunaga, L.~Shamis, J.~Kahn, X.~Ma, L.~Zettlemoyer, and O.~Levy.
\newblock Transfusion: Predict the next token and diffuse images with one multi-modal model.
\newblock {\em arXiv preprint arXiv:2408.11039}, 2024.

\bibitem{zhu2023minigpt}
D.~Zhu, J.~Chen, X.~Shen, X.~Li, and M.~Elhoseiny.
\newblock Minigpt-4: Enhancing vision-language understanding with advanced large language models.
\newblock {\em arXiv preprint arXiv:2304.10592}, 2023.

\bibitem{mmc4}
W.~Zhu, J.~Hessel, A.~Awadalla, S.~Y. Gadre, J.~Dodge, A.~Fang, Y.~Yu, L.~Schmidt, W.~Y. Wang, and Y.~Choi.
\newblock Multimodal c4: An open, billion-scale corpus of images interleaved with text.
\newblock {\em Advances in Neural Information Processing Systems}, 36, 2024.

\end{thebibliography}
}

\end{document}